\documentclass{article} 
\usepackage[table]{xcolor}
\usepackage{iclr2025_conference,times}

\usepackage{hyperref}
\usepackage{url}
\usepackage{wrapfig,lipsum,booktabs}
\usepackage{graphicx}
\usepackage{subcaption}
\usepackage{caption}
\usepackage{multirow}

\usepackage{xspace}
\usepackage{graphicx, amsmath, amssymb, caption, subcaption, multirow, overpic, textpos}
\usepackage[english]{babel}

\definecolor{citecolor}{HTML}{0071BC}
\definecolor{linkcolor}{HTML}{ED1C24}

\newlength\savewidth\newcommand\shline{\noalign{\global\savewidth\arrayrulewidth
		\global\arrayrulewidth 1pt}\hline\noalign{\global\arrayrulewidth\savewidth}}
\newcommand{\tablestyle}[2]{\setlength{\tabcolsep}{#1}\renewcommand{\arraystretch}{#2}\centering\footnotesize}
\renewcommand{\paragraph}[1]{\vspace{1.25mm}\noindent\textbf{#1}}
\newcommand\blfootnote[1]{\begingroup\renewcommand\thefootnote{}\footnote{#1}\addtocounter{footnote}{-1}\endgroup}

\newcolumntype{x}[1]{>{\centering\arraybackslash}p{#1pt}}
\newcolumntype{y}[1]{>{\raggedright\arraybackslash}p{#1pt}}
\newcolumntype{z}[1]{>{\raggedleft\arraybackslash}p{#1pt}}

\newcommand{\app}{\raise.17ex\hbox{$\scriptstyle\sim$}}

\definecolor{deemph}{gray}{0.6}

\definecolor{baselinecolor}{gray}{.9}
\newcommand{\baseline}[1]{\cellcolor{baselinecolor}{#1}}
\definecolor{Gray}{gray}{0.9}

\usepackage{pifont}
\definecolor{forestgreen}{rgb}{0.13, 0.55, 0.13}
\definecolor{brickred}{rgb}{0.8, 0.25, 0.33}

\usepackage{etoc}

\usepackage{marvosym}
\makeatletter
\def\blfootnote{\xdef\@thefnmark{}\@footnotetext}
\makeatother

\newcommand{\methodname}{Meepo}

\title{Exploring contextual modeling with linear complexity for point cloud segmentation}

\author{Yong Xien Chng$^{1,2}$\enskip
Xuchong Qiu$^{2,\dagger}$\enskip
Yizeng Han$^{1}$\enskip
Yifan Pu$^{1}$\enskip
{\bf Jiewei Cao$^{2}$\enskip
Gao Huang$^{1}$\textsuperscript{\Letter}} \\
$^1$Tsinghua University\enskip$^2$Bosch Corporate Research\\
$^{\dagger}$Project lead\enskip\textsuperscript{\Letter}Corresponding author
}

\iclrfinalcopy 
\begin{document}

\maketitle

\begin{abstract}
Point cloud segmentation is an important topic in 3D understanding. Traditionally, this task has been tackled using either the CNN or Transformer. Recently, a newcomer, Mamba, has emerged as a promising alternative, offering efficient long-range contextual modeling capabilities without the quadratic complexity associated with the Transformer’s attention mechanisms. However, despite Mamba’s potential, early efforts have \textit{all failed} to achieve better performance than the best CNN-based and Transformer-based methods. Since each of these diverse architectures possesses unique strengths and weaknesses, it's difficult to determine the most suitable approach for this task. In this work, we address this challenge by identifying the key components of an effective and efficient point cloud segmentation architecture. Specifically, we show that: 1) \textit{Spatial locality} and \textit{robust contextual understanding} are critical for strong performance, and 2) Mamba features linear computational complexity, offering superior data and inference efficiency compared to Transformers, while still being capable of delivering strong contextual understanding. Additionally, we further enhance the standard Mamba specifically for point cloud segmentation by identifying its two key shortcomings. First, the enforced causality in the original Mamba is unsuitable for processing point clouds that have no such dependencies. Second, its unidirectional scanning strategy imposes a directional bias, hampering its ability to capture the full context of unordered point clouds in a single pass. To address these issues, we carefully remove the causal convolutions and introduce a novel \textit{Strided Bidirectional SSM} to enhance the model’s capability to capture spatial relationships. Our efforts culminate in the development of a novel architecture named \textsc{\methodname}, which effectively integrates the strengths of CNN and Mamba. \textsc{\methodname} surpasses the previous state-of-the-art method, PTv3, by up to \textbf{+0.8} mIoU on key benchmark datasets, including ScanNet, ScanNet200, S3DIS, and nuScenes, while being \textbf{42.1\%} faster and \textbf{5.53$\times$} more memory efficient. Our code and data will be released.

\end{abstract}
\blfootnote{Y. X. Chng is an intern at Bosch Corporate Research.}

\section{Introduction}
Point cloud segmentation is an important topic in 3D understanding that has gained significant attention from the research community in recent years. Numerous neural network architectures have been proposed for this task, with CNN-based and Transformer-based designs being the two most prominent. Currently, Transformer-based methods consistently deliver the highest performance across numerous benchmarks. Their success is frequently credited to the attention mechanism, which can adequately capture and model context. However, the quadratic complexity of self-attention in Transformers poses a significant challenge for point cloud processing, especially when handling a large number of points. To address this, researchers have explored more efficient strategies, such as aggressive downsampling \citep{ptv1,ptv2,pointformer}, efficient attention algorithms \citep{swin3d}, and windowed attention mechanisms \citep{octformer,st,ptv3}. While these methods help reduce computational costs, they achieve this at the expense of valuable spatial and geometric information, potentially weakening the Transformer’s modeling capability and hindering its contextual understanding of the point cloud \citep{shen2021efficient}. 

Recently, State Space Models (SSMs) like Mamba \citep{jamba,mamba,moemamba,mambabyte} have emerged as a promising alternative to Transformers. These models combine aspects of recurrent neural networks \citep{gru,lstm} and convolutional neural networks \citep{cnn} within a framework rooted in classical state space theory. Similar to Transformers, SSMs provide robust contextual modeling capabilities. However, unlike Transformers, which scale quadratically with sequence length, SSMs scale linearly and maintain constant memory usage during inference. This efficiency makes SSMs particularly advantageous for tasks that require handling extensive contexts.

\begin{figure*}
\vskip -0.15in
\begin{minipage}{0.49\textwidth}
  \includegraphics[width=0.85\linewidth]{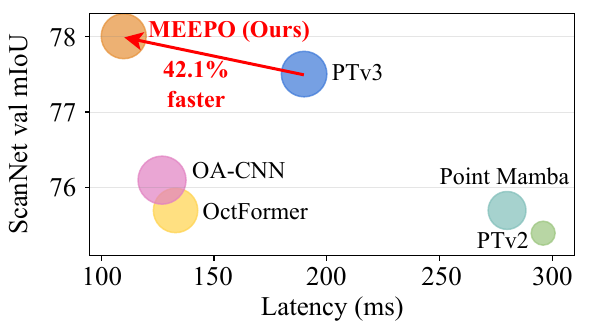}
  \captionsetup{aboveskip=-0.01pt}  
  \caption*{(a) mIoU v.s. Latency on V100}
\end{minipage}\hfill
\begin{minipage}{0.49\textwidth}
  \includegraphics[width=0.85\linewidth]{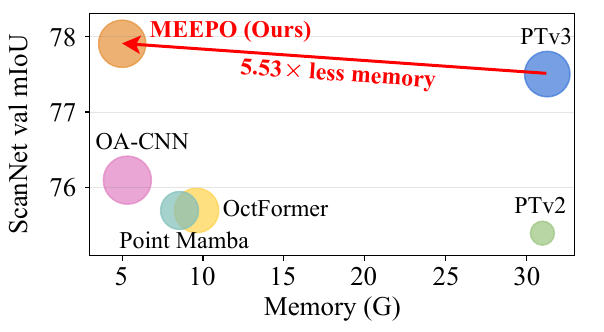}
  \captionsetup{aboveskip=-0.01pt}  
  \caption*{(b) mIoU v.s. Memory on V100}
\end{minipage}
\\
\begin{minipage}[b]{0.49\textwidth}
  \includegraphics[width=0.85\linewidth]{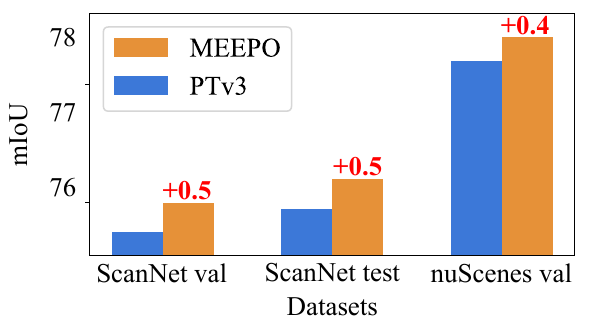}
  \captionsetup{aboveskip=-0.01pt}  
  \caption*{(c) mIoU on ScanNet and nuScenes}
\end{minipage}\hfill
\begin{minipage}[b]{0.49\textwidth}
  \includegraphics[width=0.85\linewidth]{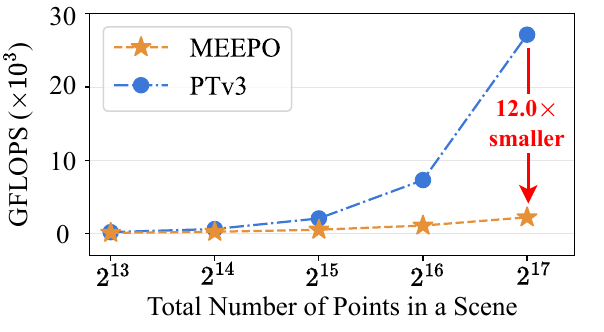}
  \captionsetup{aboveskip=-0.01pt}  
  \caption*{(d) GFLOPS v.s. Number of Points}
\end{minipage}
\caption{Performance and efficiency comparisons between our proposed method, \textsc{\methodname}, and other leading segmentation networks using the ScanNet and nuScenes dataset. By carefully combining the strengths of existing architectures, \textsc{\methodname} surpasses all previous leading methods while using much lower latency and memory. Furthermore, it easily scales to scenes with a much larger number of points and progressively improves performance as the input sequence length increases.}
\label{figure1}
\vskip -0.2in
\end{figure*}

\begin{wraptable}{r}{0.5\linewidth}
\centering
\captionsetup{aboveskip=-0.01pt}  
\captionsetup{belowskip=-10pt}  
\caption{Performance and Latency Comparison among representative Mamba-based, CNN-based and Transformer-based networks on ScanNet.}
\label{table1}
\resizebox{\linewidth}{!}{
\begin{tabular}{ccc}
    \toprule
    case & mIoU$\uparrow$ & latency$\downarrow$  \\
    \midrule
    Point Mamba \citep{pointmamba2} & 75.7 & 280 \\
    OA-CNN \citep{oacnns} & 76.1 & 141 \\
    PTv3 \citep{ptv3} (current best) & 77.5 & 183 \\
    \bottomrule
\end{tabular}}
\vskip -0.1in
\end{wraptable}

Although numerous early attempts have been made to utilize SSMs for point cloud segmentation \citep{pointmamba, pcm, pointmamba2}, these efforts have been largely \textit{discouraging}, as their performance significantly lags behind that of leading CNN-based and Transformer-based models. As illustrated in Tab.~\ref{table1}, the recently proposed Point Mamba network \citep{pointmamba2} not only incurs roughly twice the latency of the CNN-based OA-CNN \citep{oacnns} and Transformer-based PTv3 \citep{ptv3} but also underperforms them by \textbf{0.4} points and \textbf{2.2} points in mIoU on the ScanNet \citep{scannet} validation dataset, respectively. These shortcomings underscore the ongoing debate regarding the optimal architecture for point cloud segmentation, leading to the pivotal question: \textit{What constitutes an efficient and effective model architecture for point cloud segmentation?}

In this work, we aim to provide valuable insights for the design of point cloud segmentation architectures. To achieve this, we conduct a preliminary analysis in Sec.~\ref{section2} to examine the properties of the three most popular architectures, namely the CNN-based, Transformer-based and Mamba-based networks. Using the meta-architecture presented in Fig.~\ref{figure2}, which seamlessly integrates different operators from the CNN, Transformer and Mamba architectures, we compare these architectures across three key dimensions: \textit{contextual understanding capability}, \textit{local sensitivity}, and \textit{network efficiency}. Our analysis reveals that while CNNs excel at local modeling, they lack the ability to capture broader context. Transformers can adequately capture contextual information but are inefficient due to unnecessary long-range attention and quadratic computational complexity. Mambas strike a balance by efficiently providing essential contextual understanding with linear complexity. Given the distinct strengths and limitations of each architecture, we hypothesize that an integrated architecture combining their best features could potentially yield a more powerful and efficient model.

Building on the insights gained, we systematically explore various block choices, placements, and quantities within the previously proposed meta-architecture to determine the optimal arrangement. Through this process, we identify the CNN-Mamba block as the optimal elementary block for our architecture. As depicted in Fig.~\ref{figure6}(a), the CNN-Mamba block comprises a sequential stack of sparse convolution layers and Mamba modules. Notably, the Attention module is ultimately excluded from the final architecture because the more efficient Mamba module can already provide sufficient contextual understanding for this task.

With the macro-level design established, we turn our attention to the micro-level design, specifically assessing whether the standard Mamba module, originally designed for sequential processing, is suitable for point cloud segmentation. Our investigation shows that it is not. In particular, we identify two major shortcomings of the standard Mamba module when applied to this task:

\begin{enumerate} 
\vspace{-1mm}
\item \textbf{Loss of spatial information due to enforced causality}: Mamba's use of causal convolutions introduces unnecessary artificial dependencies that can disrupt the inherent spatial relationships in point cloud data, ultimately reducing its effectiveness for point cloud data.
\item \textbf{Directional bias due to unidirectional scan}: Mamba's unidirectional scan strategy inherently favors certain data points over others, creating a directional bias. This bias undermines its ability to fully grasp the context of unordered point clouds in a single pass.
\end{enumerate}

To address these issues, we propose two corresponding improvements to the standard Mamba module. Specifically, we 1) remove the causality constraint, and 2) incorporate a novel \textit{Strided Bidirectional SSM} to enhance its context and spatial understanding.

Our work culminates in \textsc{\methodname}, a novel point cloud segmentation architecture which seamlessly integrates Mamba's efficiency and strong contextual understanding with the local sensitivity of CNN. As shown in Fig.~\ref{figure1}, \textsc{\methodname} not only achieves significantly lower inference latency but also surpasses the performance of other leading segmentation models across both indoor and outdoor scenes. Specifically, it outperforms the previous best method, PTv3, by up to \textbf{+0.8} mIoU across ScanNet, ScanNet200, S3DIS, and nuScenes datasets, with \textbf{42.1\%} smaller latency and \textbf{5.53}$\times$ smaller memory usage. \textsc{\methodname} also scales much better with respect to the size of point cloud. As shown in Fig.~\ref{figure1}(d), it achieves up to $12\times$ fewer GFLOPs than PTv3 when processing a scene with $2^{17}$ ($131,072$) points.

In summary, the contributions of our paper are as follows:
\vspace{-1mm}
\begin{enumerate} 
\item We carefully analyze existing point cloud segmentation networks, identifying the importance of \textit{spatial locality} and \textit{robust contextual understanding} in achieving high performance. We then reveal that Transformer's global attention is unnecessary and inefficient for this task as Mambas offer comparable contextual modeling with linear complexity.
\item We propose two corresponding solutions to address the limitations of Mamba when applied to point cloud segmentation, namely the removal of the causality constraint and the incorporation of an innovative \textit{Strided Bidirectional SSM} to enhance contextual understanding.
\item We introduce \textsc{\methodname}, a novel architecture that solely utilizes the efficient sparse convolution to provide spatial locality and the efficient SSM to provide robust contextual understanding. \textsc{\methodname} not only consistently outperforms previous best method, PTv3 across multiple key benchmark datasets, but is also much faster and much more memory efficient.
\end{enumerate}

\vspace{-2mm}
\section{Preliminaries}
In this section, we briefly introduce state space model (SSM) to facilitate subsequent discussion.

\textbf{SSM's formulation. } SSM is a type of sequence model \citep{mamba} that maps an input sequence $x(t) \in \mathbb{R}$ to an output sequence $y(t) \in \mathbb{R}$ through a latent state $h(t) \in \mathbb{R}^N$:
\begin{equation}
h'(t) = \mathbf{A} h(t) + \mathbf{B}\mathbf{x}(t), \quad y(t) = \mathbf{C} h(t).
\end{equation}
Using zero-order hold (ZOH) rule, we can discretize the continuous parameters $(\mathbf{\Delta}, \mathbf{A}, \mathbf{B})$ as:
\begin{equation}
\bar{\mathbf{A}} = \exp(\Delta \mathbf{A}), \quad \bar{\mathbf{B}} = (\Delta \mathbf{A})^{-1}(\exp(\Delta \mathbf{A}) - \mathbf{I}) \cdot \Delta \mathbf{B},
\end{equation}
where $\Delta$ is the discretization step size and $\mathbf{I}$ is the identity matrix. Under this discretization rule, the hidden state $h_t$ can be computed efficiently as a linear recurrence:
\begin{equation}
h_t = \bar{\mathbf{A}} h_{t-1} + \bar{\mathbf{B}} x_t, \quad y_t = \mathbf{C} h_t.
\end{equation}

Mamba is a recent popular SSM model that sets the SSM parameters to be functions of the input:
\begin{equation}
\mathbf{B}_t = f_B(x(t)) \quad \mathbf{C}_t = f_C(x(t)) \quad \mathbf{\Delta}_t = f_\Delta(x(t))
\end{equation}
This allows the model to selectively propagate or discard information based on the input. In practice, matrix $\mathbf{A}$ is typically set as a diagonal matrix, ensuring that all elements of $\bar{\mathbf{A}} = \exp(\Delta \mathbf{A})$ lie between 0 and 1. Consequently, $\bar{\mathbf{A}}$ can be viewed as a \textit{forget gate}, which controls how much information from the previous hidden state, $\textit{h}_t$, is retained.


\section{Analysis}
\label{section2}

\begin{figure}[t]
  \vskip -0.15in
    \centering
    \includegraphics[width=0.9\linewidth]{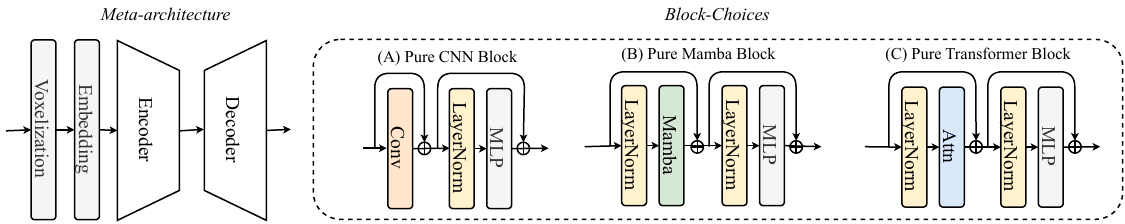} 
    \caption{Proposed meta-architecture and various block options used for analysis. The model that exclusively uses choice A is called \textit{Pure CNN}, the model that exclusively uses choice B is called \textit{Pure Mamba}, and the model that exclusively uses choice C is called \textit{Pure Transformer}.}
    \label{figure2}
\end{figure}

In this section, we delve into the key requirements for effective point cloud segmentation, showing the importance of capturing both local and contextual features via model architecture. Robust local modeling is essential for maintaining point-level consistency, especially when object boundaries are unclear, while strong contextual modeling is crucial for identifying occluded or ambiguously shaped objects. With these needs in mind, we evaluate various architectures, assessing their strengths and weaknesses in providing these important properties for effective point cloud segmentation.

\paragraph{Meta-architecture for analysis.} To facilitate our analysis, we first propose a meta-architecture for point cloud segmentation. This meta-architecture follows an encoder-decoder framework, featuring a 5-stage encoder and a 4-stage decoder. Following previous works \citep{pcm,pointmamba2}, the input points are first voxelized into non-overlapping segments and arranged into an ordered sequence using a locality-preserving space-filling curve \citep{morton,spacefilling}. These voxels are then processed by an embedding module that employs a single submanifold sparse convolution layer \citep{submanifold}. GridPooling and GridUnpooling operations are applied at the end and beginning of each encoding stage, respectively, to downsample and upsample the point cloud \citep{ptv3}. Within this meta-architecture, we train three similarly sized models, each using either the CNN, Mamba, or Transformer blocks, as shown in Fig.~\ref{figure2}, to evaluate their performance both qualitatively and quantitatively. These models are referred to as \textit{Pure CNN}, \textit{Pure Mamba}, and \textit{Pure Transformer}, respectively. We hypothesize that the global attention of \textit{Pure Transformer} and the long-range sequential modeling capabilities of \textit{Pure Mamba} are well-suited for contextual modeling. \textit{Pure Mamba}, in particular, offers additional advantages such as greater inference efficiency due to its linear complexity and improved data efficiency through inductive bias provided by the forget gate. In contrast, \textit{Pure CNN} excels at local modeling, leveraging its inherent spatial locality. 

\begin{wraptable}{r}{0.38\linewidth}
\centering
\captionsetup{aboveskip=2pt}  
\captionsetup{belowskip=-10pt}  
\caption{PTv3 without spatial locality perform much worse on ScanNet val.}
\label{table2}
\resizebox{0.95\linewidth}{!}{
\begin{tabular}{lc}
    \toprule
    Case & mIoU$\uparrow$  \\
    \midrule
    PTv3 \citep{ptv3} & 77.5 \\
    \quad Remove spatial locality & 69.3 (\textcolor{brickred}{\textbf{-8.2}}) \\
    \quad Remove contextual modeling & 73.2 (\textcolor{brickred}{\textbf{-4.3}}) \\
    \bottomrule
\end{tabular}
}
\vskip -0.08in
\end{wraptable}
\textbf{1. Are both local and contextual modeling important for point cloud segmentation?} To assess the individual contributions of local and contextual modeling in point cloud segmentation, we analyze the performance impact of removing key components from the current leading network, PTv3 \citep{ptv3}. Specifically, we investigate the effects of eliminating the sparse convolution layers, which capture spatial locality, and attention modules, which model contextual relationships. As detailed in Tab.~\ref{table2}, removing the sparse convolution layers results in a substantial performance drop of \textbf{-8.2} mIoU, while removing the attention modules leads to a decrease of \textbf{-4.3} mIoU. These results highlight the critical importance of both local and contextual modeling in achieving effective point cloud segmentation.

\begin{figure}[t]
    \centering
    \begin{minipage}[b]{0.32\textwidth}
        \centering
        \includegraphics[width=0.9\linewidth]{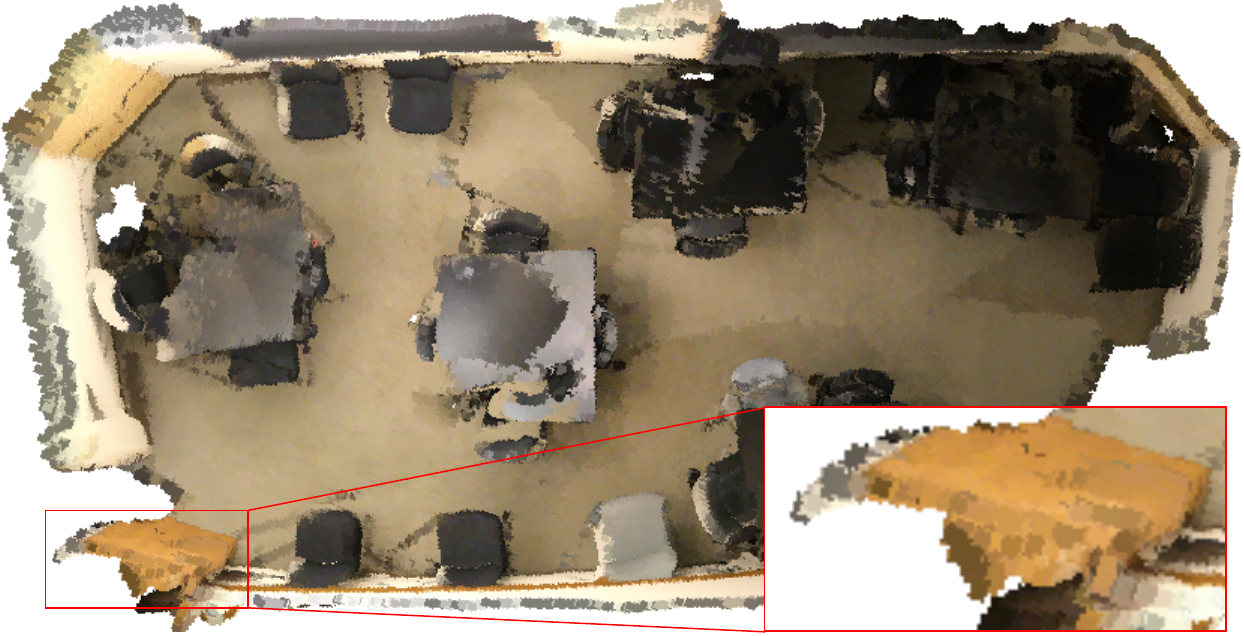}
        \captionsetup{aboveskip=0pt}  
        \captionsetup{belowskip=-5pt}  
        \caption*{Image}
    \end{minipage}
    \hfill
    \begin{minipage}[b]{0.32\textwidth}
        \centering
        \includegraphics[width=0.9\linewidth]{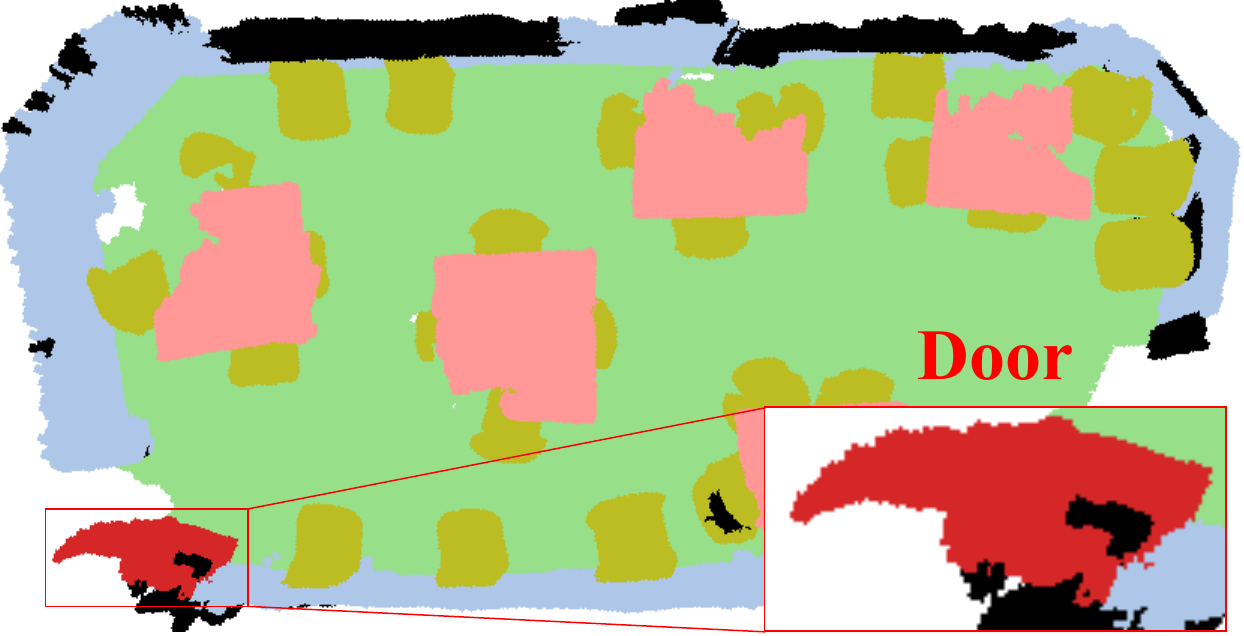}
        \captionsetup{aboveskip=0pt}  
        \captionsetup{belowskip=-5pt} 
        \caption*{Ground Truth}
    \end{minipage}
    \hfill
    \begin{minipage}[b]{0.32\textwidth}
        \centering
        \includegraphics[width=0.9\linewidth]{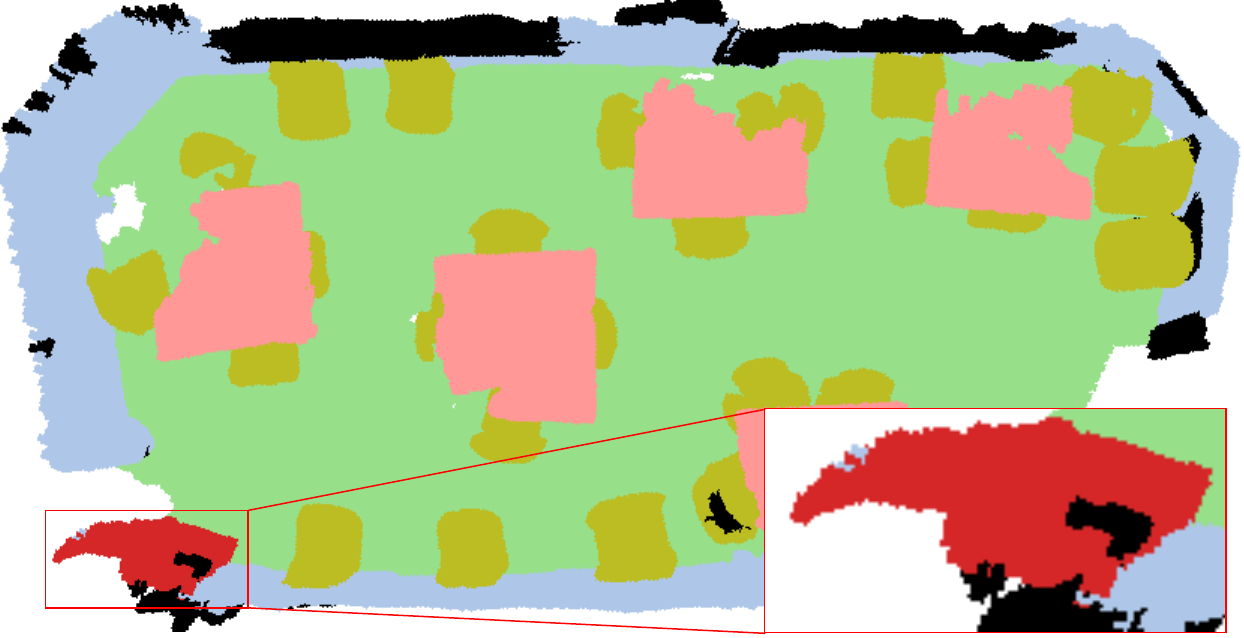}
                \captionsetup{aboveskip=0pt}  
        \captionsetup{belowskip=-5pt} 
        \caption*{\textsc{\methodname}'s Prediction}
    \end{minipage}
    \vskip\baselineskip
    \begin{minipage}[b]{0.32\textwidth}
        \centering
        \includegraphics[width=0.9\linewidth]{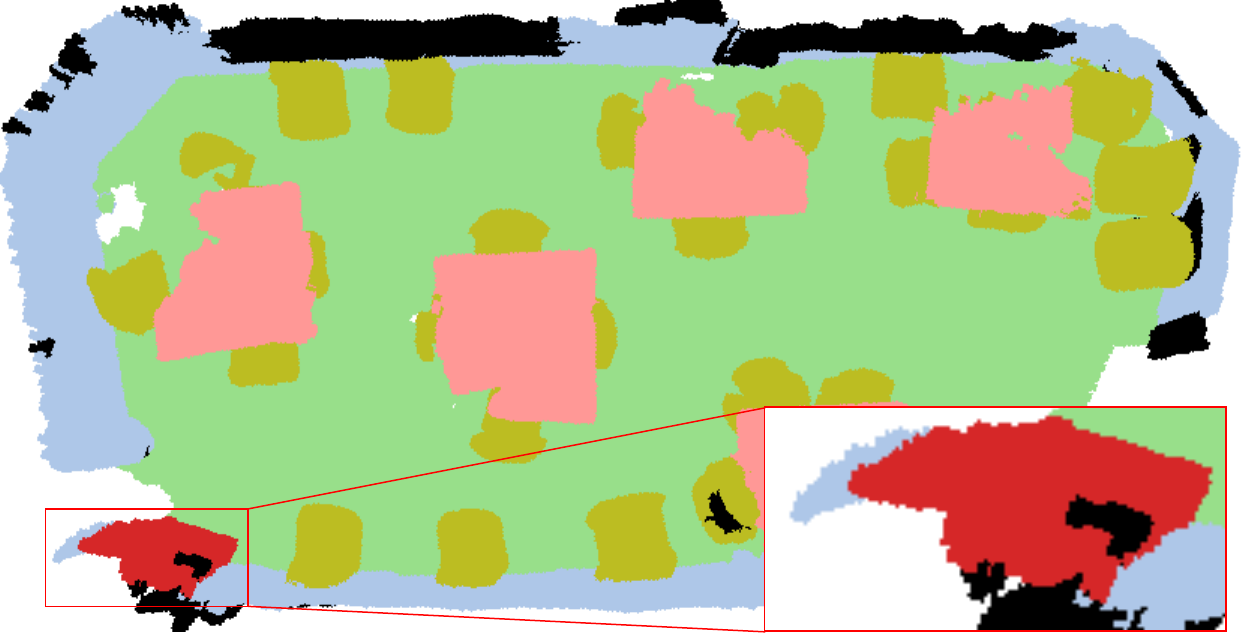}
                \captionsetup{aboveskip=0pt}  
        \captionsetup{belowskip=-5pt} 
        \caption*{Transformer's Prediction}
    \end{minipage}
    \hfill
    \begin{minipage}[b]{0.32\textwidth}
        \centering
        \includegraphics[width=0.9\linewidth]{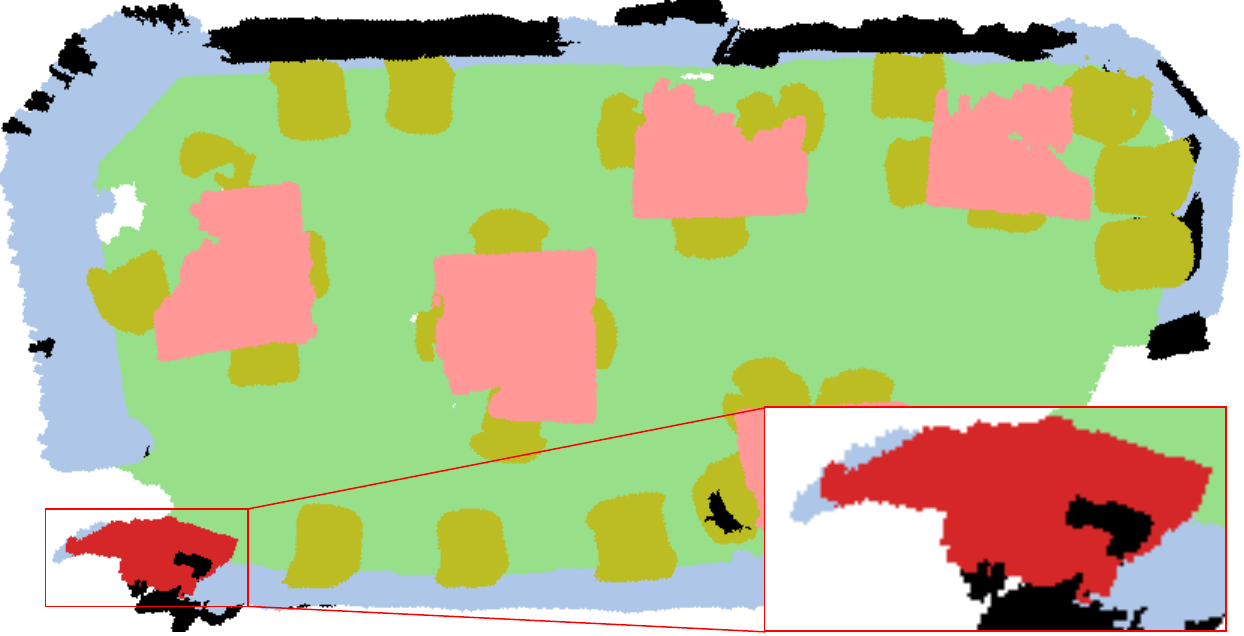}
                \captionsetup{aboveskip=0pt}  
        \captionsetup{belowskip=-5pt} 
        \caption*{Mamba's Prediction}
    \end{minipage}
    \hfill
    \begin{minipage}[b]{0.32\textwidth}
        \centering
        \includegraphics[width=0.9\linewidth]{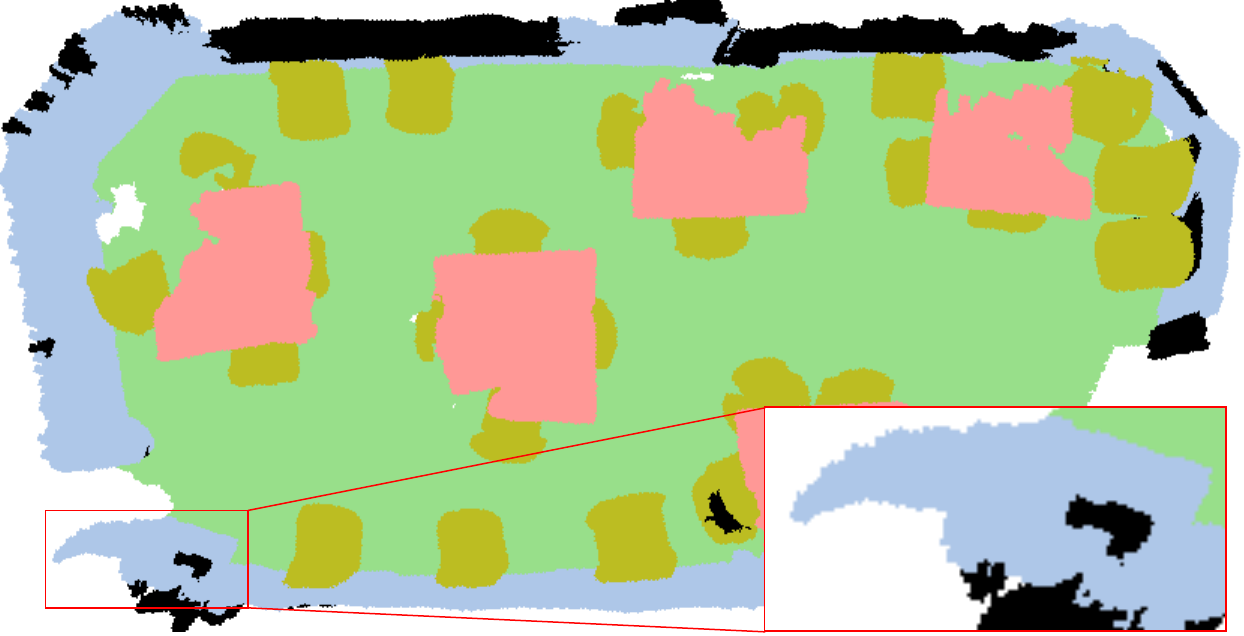}
                \captionsetup{aboveskip=0pt}  
        \captionsetup{belowskip=-5pt} 
        \caption*{CNN's Prediction}
    \end{minipage}
    \caption{Both Transformer and Mamba models incorporate mechanisms to learn long-range dependencies, allowing them to accurately interpret occluded regions and areas with similar textures.}
    \label{figure3}
    \vskip -0.1in
\end{figure}

\textbf{2. Which model is more effective at providing contextual understanding?} To investigate this, we perform another experiment using PTv3, by modifying its full attention mechanism to a window-based approach. By varying the window size, we control the amount of context the model processes. \begin{wrapfigure}{r}{0.4\textwidth}
\vspace{-0.08in} 
  \begin{center}
    \includegraphics[width=\linewidth]{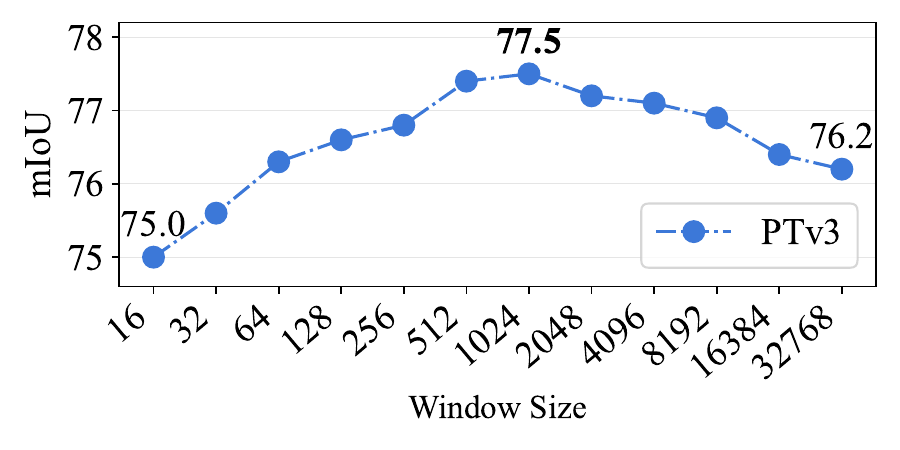}
  \end{center}
  \captionsetup{aboveskip=-5pt}  
    \captionsetup{belowskip=-5pt}  
  \caption{Analysis of a representative Transformer-based model, PTv3, demonstrates that additional context beyond a certain amount is unnecessary.}
  \label{figure4}
\end{wrapfigure}
As shown in Fig.~\ref{figure4}, increasing the window size progressively improves performance from 24 to 1024, peaking at 1024 before gradually declining as the window size grows further. This indicates that while contextual understanding is beneficial, too much context can become detrimental, underscoring the importance of finding a balance between local and global modeling. Meanwhile, Mamba emerges as a strong candidate due to its emphasis on local processing while still capturing broader context when necessary. To provide evidence for this, we visualize the performance of different models using a representative example in Fig.~\ref{figure3}, which depicts an office room with a challenging-to-discern door in the bottom left corner. Try to focus on the boxed region in each image and identify the object category. This task is challenging because the door’s point cloud is partially occluded and its texture and color closely match the surrounding wall and floor, requiring a comprehensive understanding of the scene. In this scenario, where contextual understanding is crucial, Mamba demonstrates superior performance by effectively balancing local and global context. It outperforms the Transformer model, which in turn surpasses the CNN, thereby confirming Mamba's efficacy in contextual understanding.

\textbf{3. Which model is more effective at providing spatial locality?} To compare the performance of the aforementioned model architectures in this regard, we visualize their predictions in Fig.~\ref{figure4} using another insightful example of a room with a centrally placed table whose boundaries are poorly defined due to severe overexposure. Without zooming in, the boundaries are hard to discern, requiring strong local feature extraction capabilities. In this challenging scenario, the results support our hypothesis regarding spatial locality. The CNN, equipped with sparse convolutional layers specifically designed to capture local spatial patterns, outperforms both Mamba and the Transformer. Mamba exhibits slight spillage of floor pixels onto the table, while the Transformer mislabels a significant portion of the table area. These results demonstrate the CNN's superior capability in local modeling.


\begin{figure}[t]
    \vskip -0.15in
    \centering
    \begin{minipage}[b]{0.32\textwidth}
        \centering
        \includegraphics[width=0.85\linewidth]{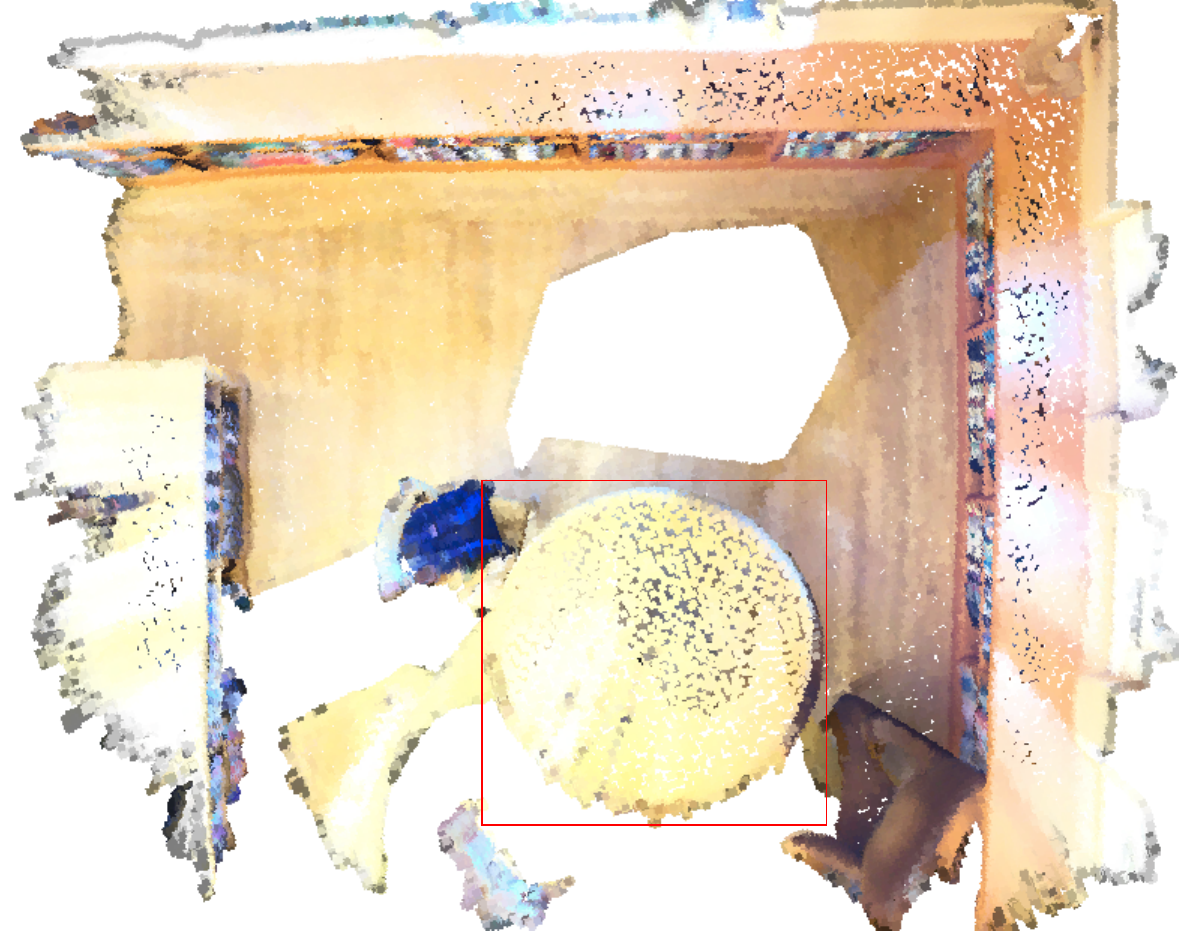}
                \captionsetup{aboveskip=0pt}  
        \captionsetup{belowskip=-5pt}  
        \caption*{Image}
    \end{minipage}
    \hfill
    \begin{minipage}[b]{0.32\textwidth}
        \centering
        \includegraphics[width=0.85\linewidth]{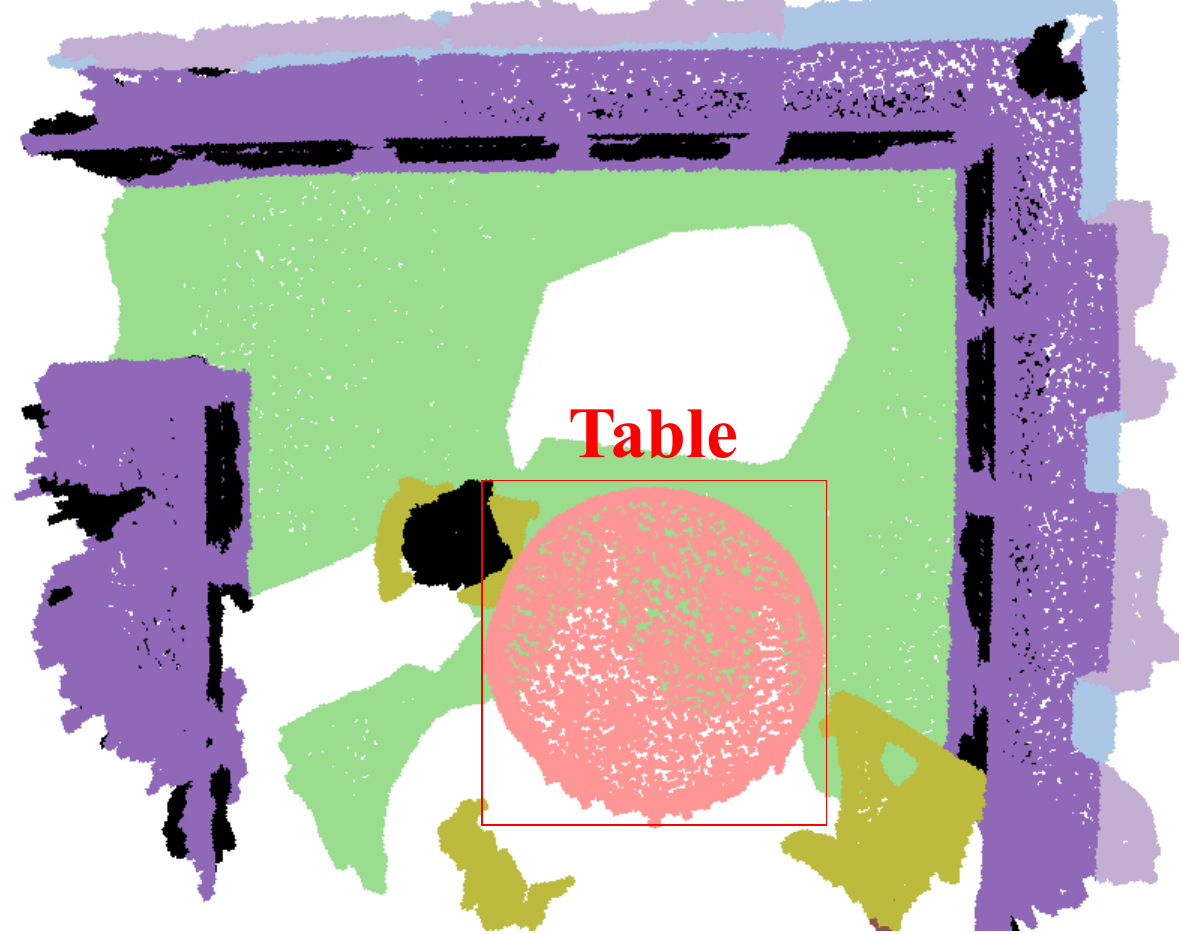}
                \captionsetup{aboveskip=0pt}  
        \captionsetup{belowskip=-5pt}  
        \caption*{Ground Truth}
    \end{minipage}
    \hfill
    \begin{minipage}[b]{0.32\textwidth}
        \centering
        \includegraphics[width=0.85\linewidth]{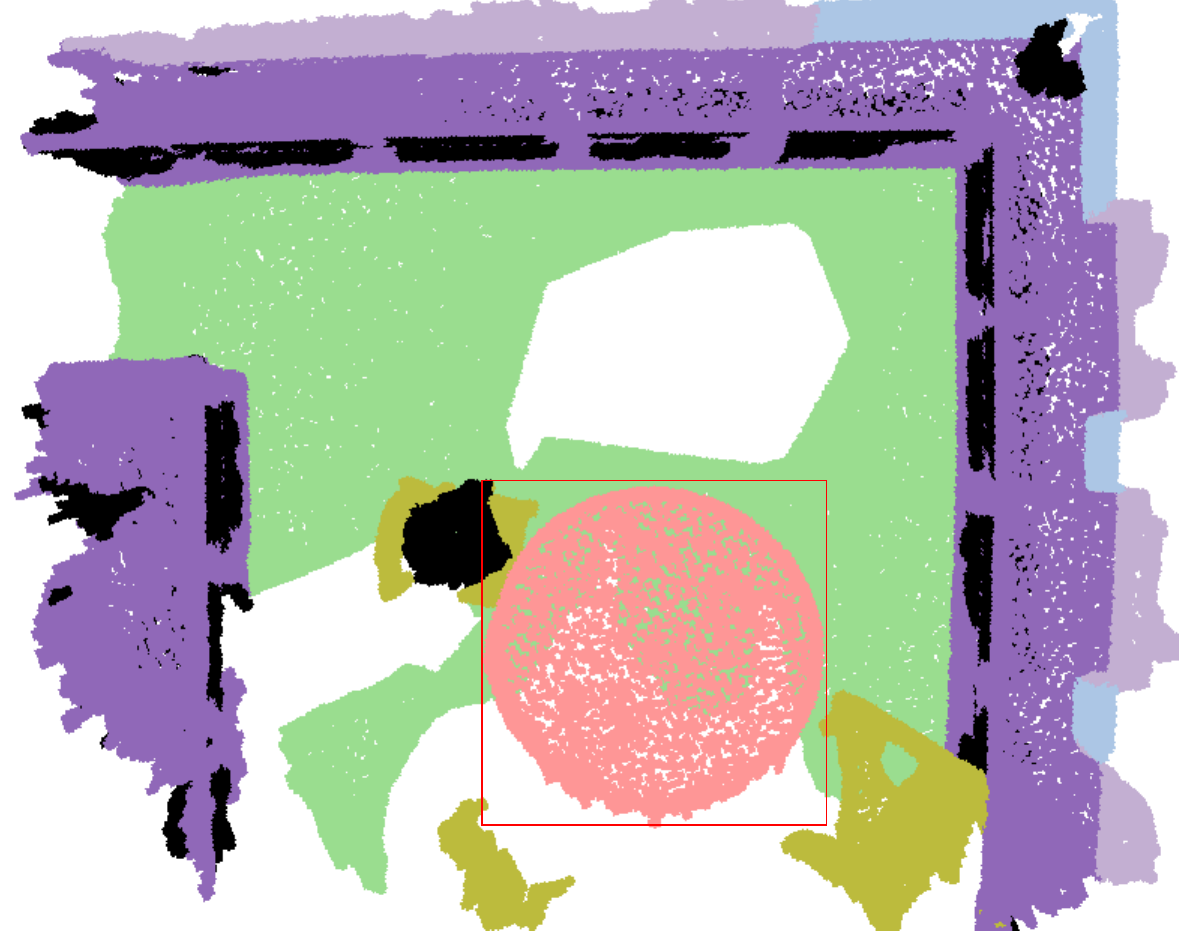}
                \captionsetup{aboveskip=0pt}  
        \captionsetup{belowskip=-5pt}  
        \caption*{\textsc{\methodname}'s Prediction}
    \end{minipage}
    \vskip\baselineskip
    \begin{minipage}[b]{0.32\textwidth}
        \centering
        \includegraphics[width=0.85\linewidth]{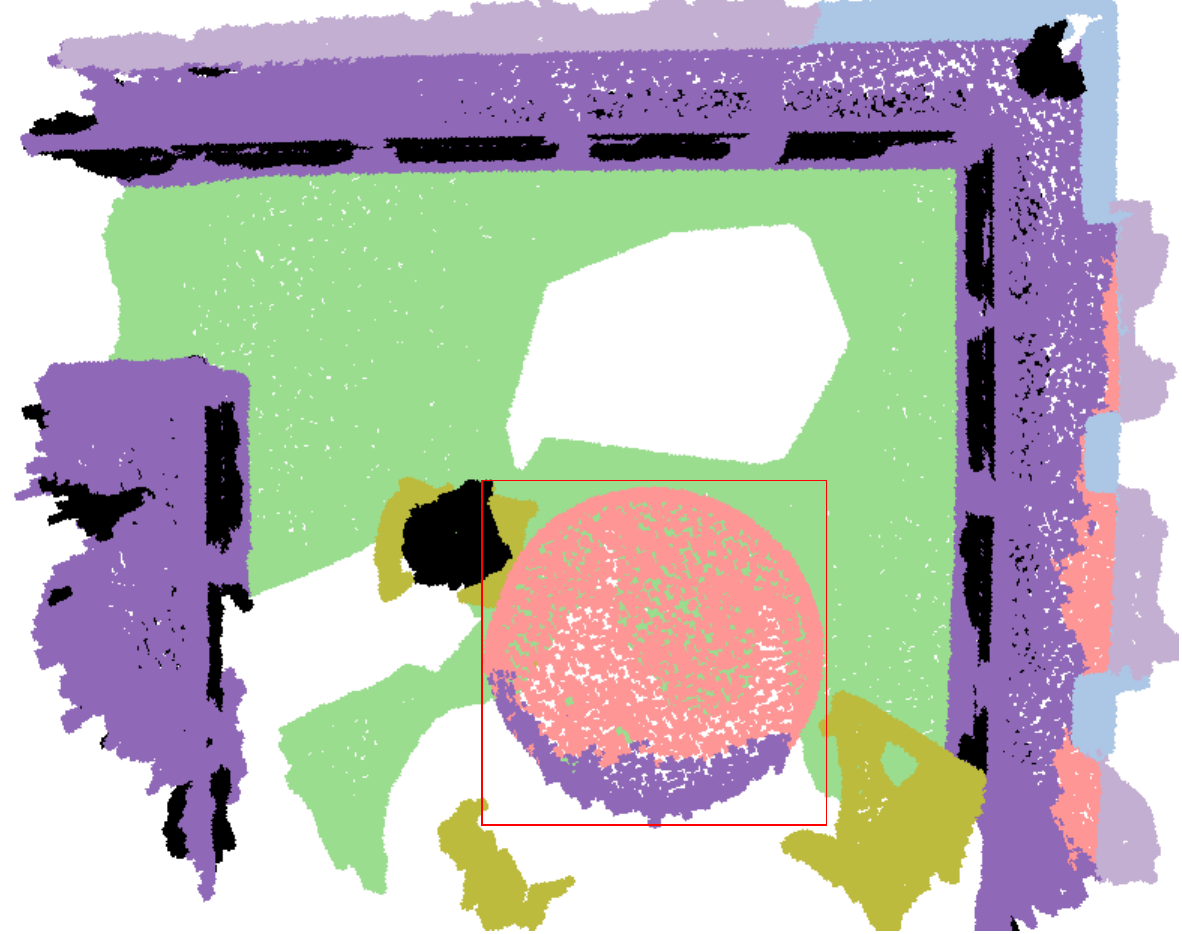}
                \captionsetup{aboveskip=0pt}  
        \captionsetup{belowskip=-5pt}  
        \caption*{Transformer's Prediction}
    \end{minipage}
    \hfill
    \begin{minipage}[b]{0.32\textwidth}
        \centering
        \includegraphics[width=0.85\linewidth]{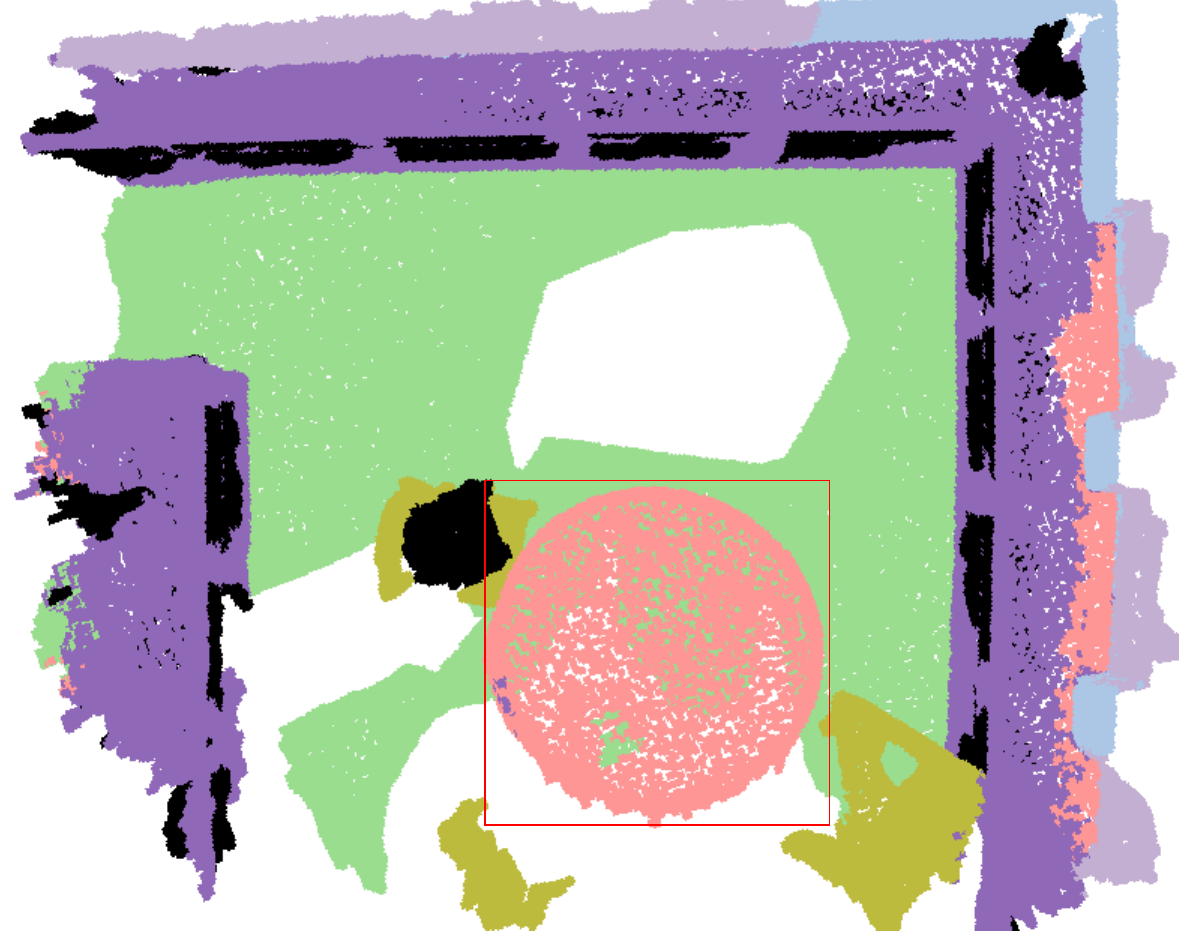}
                \captionsetup{aboveskip=0pt}  
        \captionsetup{belowskip=-5pt}  
        \caption*{Mamba's Prediction}
    \end{minipage}
    \hfill
    \begin{minipage}[b]{0.32\textwidth}
        \centering
        \includegraphics[width=0.85\linewidth]{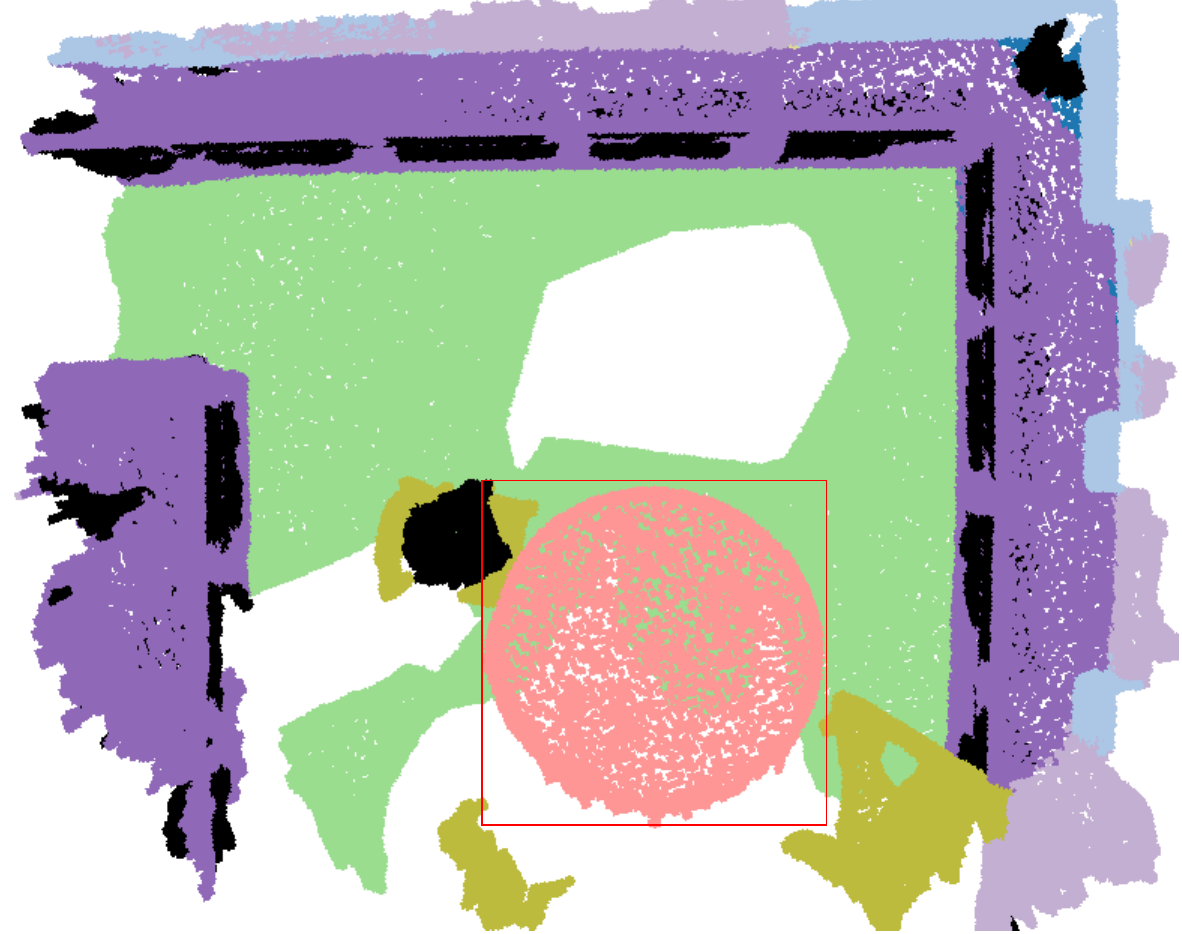}
                \captionsetup{aboveskip=0pt}  
        \captionsetup{belowskip=-5pt}  
        \caption*{CNN's Prediction}
    \end{minipage}
    \caption{Comparison of model performance on tasks requiring robust local modeling. CNNs excel due to spatial convolutions, and Mambas benefit from its locally-biased forget gate. Transformers, lacking specialized local modeling mechanisms, often produce inaccurate predictions.}
    \label{figure5}
    \vskip -0.08in
\end{figure}

\textbf{4. Which model architecture is more efficient?} Let $C$, $C_{\text{in}}$, and $C_{\text{out}}$ represent the channel sizes, $N$ the SSM's state dimension, $E$ the SSM's expansion factor, $L$ the number of points or input sequence length, $K$ the depthwise convolution kernel size in Mamba and $k$ the convolution kernel size in sparse convolutions within CNN, the computational complexities of the core operations in the Transformer, Mamba, and CNN models are as follows:
\small
\begin{align}
& \Omega(\text{Transformer}) & = & \hspace{4mm} \underbrace{4 \cdot L \cdot C^2}_{\text{qkv and output projections}} & + & \hspace{4mm} \underbrace{2 \cdot L^2 \cdot C}_{\text{attention}}          &     &     & = & \hspace{2mm} O(L^2), \\
& \hspace{3mm} \Omega(\text{Mamba})      & = & \hspace{7mm} \underbrace{9 \cdot L \cdot C \cdot N}_{\text{SSM}}                                   & + & \hspace{2mm} \underbrace{L \cdot C \cdot K}_{\text{depthwise conv1d}}   & + & \underbrace{3 \cdot L \cdot C^2 \cdot E}_{\text{input and output projections}} & = & \hspace{2mm} O(L), \\
& \hspace{5mm} \Omega(\text{CNN})        & = & \hspace{3mm} \underbrace{2 \cdot C_{\text{in}} \cdot C_{\text{out}} \cdot k^3 \cdot L}_{\text{convolution}} & + & \hspace{2mm} \underbrace{L \cdot C_{\text{in}} \cdot C_{\text{out}}}_{\text{bias addition}} &     &     & = & \hspace{2mm} O(L).
\end{align}
\normalsize
Both Mamba models and CNNs scale linearly with respect to \( L \), while Transformers scale quadratically. Consequently, for point cloud processing tasks involving hundreds of thousands of points, Transformers can be significantly slower. When comparing CNNs and Mamba models of similar sizes, CNNs are generally faster in practice. This is because CNNs typically have fewer layers when the total number of parameters is the same, as each 3D sparse convolution layer in a CNN contains significantly more parameters than a corresponding block in a Mamba model.

\textbf{Key insights:} Point cloud segmentation requires both effective local modeling and a comprehensive understanding of contextual information. While CNNs are effective for local modeling, contextual modeling demands a different approach. Although Transformers can handle contextual information, they are inefficient because very long-range attention is unnecessary, e.g., PTv3 performance peaks at a window size of 1024. Mambas, on the other hand, strike a balance between local and global modeling, providing the necessary contextual understanding with linear complexity instead of the quadratic complexity associated with Transformers. Given the distinct strengths and limitations of each architecture, it's essential to explore models that can holistically integrate their strengths.

\section{Method}

Building on insights from previous analysis, we introduce \textsc{\methodname}, a novel \underline{m}odel that is both \underline{e}ffective and \underline{e}fficient for \underline{po}int cloud segmentation. \textsc{\methodname} adopts the same meta-architecture presented in Fig.~\ref{figure2} but incorporates the CNN-Mamba block as a core component to facilitate the seamless integration of local and contextual modeling. Additionally, we introduce two micro-level modifications to the standard Mamba module to address its limitations in point cloud segmentation.

\subsection{Macro Architecture Design}
\begin{figure}[t]
  \vskip -0.16in
    \centering
    \includegraphics[width=0.9\linewidth]{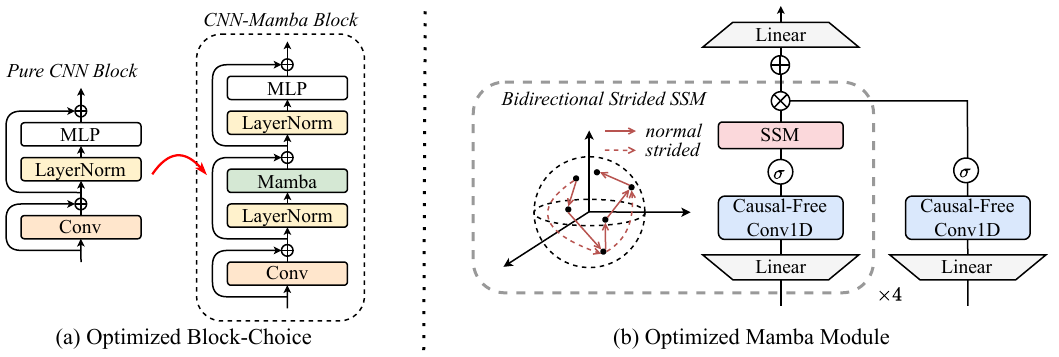} 
      \captionsetup{aboveskip=5pt}  
    \caption{Our proposed architecture, \textsc{\methodname}, integrates CNN-Mamba blocks throughout the proposed meta-architecture to harness their strengths in local and contextual modeling. To optimize for point cloud segmentation, \textsc{\methodname} modifies the standard Mamba by replacing causal convolutions with regular convolutions, preserving critical spatial information. Additionally, it introduces a novel \textit{Bidirectional Strided SSM}, which enhances contextual modeling by minimizing directional bias.}
    \label{figure6}
\end{figure}

\noindent \textbf{Optimizing block choices for  point cloud segmentation.} To identify the most effective block configuration for integrating local and contextual modeling, we draw inspiration from the widely adopted sequential combination of local convolutional layers and contextual operators in previous point cloud segmentation studies \citep{octformer, ptv3} and evaluate two possible candidates: the CNN-Mamba block and the CNN-Transformer block. As depicted in Fig.~\ref{figure6}, the CNN-Mamba block comprises a sparse convolution layer, followed by a Mamba module and an MLP layer. In contrast, the CNN-Transformer block follows the same structure but substitutes the Mamba module with an Attention module. Our comprehensive ablation study, presented in Tab.~\ref{table7}(b), demonstrates that replacing CNN-Transformer blocks with CNN-Mamba blocks throughout the network (22 blocks in total) achieves the best performance and efficiency. These findings validate our hypothesis that the Mamba module offers superior contextual understanding while maintaining efficiency.

\subsection{Micro Architecture Design}

\textbf{Optimizing Mamba for point cloud segmentation.} Despite its impressive speed and performance, \textsc{\methodname} without micro-level optimizations still does not surpass the leading point cloud segmentation network, PTv3, in accuracy. This aligns with previous research, which consistently demonstrates that attempts to apply Mamba have all failed to outperform well-established models for this task \citep{pcm,pointmamba2}. A closer analysis reveals that the standard Mamba, originally designed for sequential data, is inherently unsuited for processing unordered point clouds. We identify two key limitations in the standard Mamba and propose the solutions to overcome them.

\textbf{1. Loss of spatial information due to enforced causality:} Mamba was originally designed for sequential data with clear causal relationships. However, point clouds lack such dependencies, as their spatial relationships are multidimensional, requiring simultaneous consideration of points holistically. The causal convolutions in standard Mamba impose a causal dependency that disrupts these essential spatial relationships, making it ineffective for handling spatial data like point clouds.

\textbf{Proposed Solution:} \textit{Causal-Free Mamba.} To address this issue, we propose replacing the causal convolution with a standard convolution, resulting in the \textit{Causal-Free Mamba} module, as illustrated in Fig.\ref{figure6}(b). This modification eliminates the limitations of the original Mamba architecture when processing spatial data, greatly enhancing its performance in point cloud segmentation tasks.

\textbf{2. Directional bias due to unidirectional scan:} Mamba's unidirectional scanning method introduces a directional bias in the representation learning of point clouds because some parts are scanned first while others are scanned later. Such sequential processing is ill-suited for orderless point cloud data, where all information should be treated equally. This approach can lead to models prioritizing information from later stages of the scan, overlooking details captured earlier. The bias is further amplified by factors such as noise, occlusions, or reduced data density at the beginning of the scan, which can degrade the quality of the early data. As a result, important features captured early in the scan may be missed or inaccurately interpreted, leading to segmentation errors where boundaries are poorly defined, and features are incorrectly merged or omitted.

\textbf{Proposed Solution:} \textit{Bidirectional Strided SSM.} To address this issue, we propose an innovative multi-directional scanning approach to reduce the directional bias of Mamba. Unlike the standard Mamba, which employs a unidirectional scan, this scanning method processes data in four distinct scanning directions: forward, backward, \(n\)-strided forward, and \(n\)-strided backward. In an \(n\)-strided forward scan, every \(n\)-th token is skipped, and the scan pattern restarts from the beginning once the end is reached. For example, given the sequence 1, 2, 3, 4, 5, 6, a 2-strided forward scan would process it as 1, 3, 5, 2, 4, 6. The backward scan operates similarly but in reverse order. This multi-directional scanning approach can effectively expand Mamba's receptive field, reduce information loss, and improve overall performance by shortening the information flow path.

\section{Experiments}

\begin{table}[t!]
  \vskip -0.1in
      \begin{minipage}{.59\linewidth}
        \centering
        \captionsetup{aboveskip=-0.01pt}  
        \caption{Indoor semantic segmentation comparison \\on ScanNet, ScanNet200, S3DIS Area 5.}
        \label{table3}
        \resizebox{0.98\linewidth}{!}{\begin{tabular}{cccccc}
            \toprule
            \multirow{2}{*}{Method} & \multicolumn{2}{c}{ScanNet} & \multicolumn{2}{c}{ScanNet200} & S3DIS \\
            \cmidrule(lr){2-3} \cmidrule(lr){4-5} \cmidrule(lr){6-6}
            & Val & Test & Val & Test & Area5\\
            \midrule
            PCM \citep{pcm} & - & - & - & - & 63.4 \\
            PointNeXt \citep{pointnext} & 71.5 & 71.2 & - & - & 70.5 \\
            MinkUNet \citep{minkunet} & 72.2 & 73.6 & 25.0 & 25.3 & 65.4 \\
            ST \citep{st} & 74.3 & 73.7 & - & - & 72.0 \\
            PTv2 \citep{ptv2} & 75.4 & 74.2 & 30.2 & - & 71.6  \\
            OctFormer \citep{octformer} & 75.7 & 76.6 & 32.6 & 32.6 & - \\
            Point Mamba \citep{pointmamba2} & 75.7 & - & - & - & - \\
            OA-CNNs \citep{oacnns} & 76.1 & 75.6 & 32.3 & 33.3 &  71.1 \\
            Swin3D \citep{swin3d} & 76.4 & - & - & - & 72.5 \\
            PTv3 \citep{ptv3} & 77.5 & 77.9 & 35.2 & 37.8 &  73.4  \\
            \midrule
            \baseline{\textsc{\methodname (Ours)}} & \baseline{\textbf{78.0}} & \baseline{\textbf{78.4}} &  \baseline{\textbf{36.0}} & \baseline{\textbf{38.5}} & \baseline{\textbf{73.5}} \\
            \bottomrule
        \end{tabular}}
        \end{minipage}
        \hfill
        \begin{minipage}{.4\linewidth}
        \centering
        \captionsetup{aboveskip=-0.01pt}  
        \caption{Outdoor semantic segmentation comparison on nuScenes.} 
        \label{table4}
        \resizebox{0.98\linewidth}{!}{\begin{tabular}{ccc}
            \toprule
            \multirow{2}{*}{Method} & \multicolumn{2}{c}{nuScenes}  \\
            \cmidrule(lr){2-3} 
            & val & test \\
            \midrule
            AF2S3Net \citep{af2s3net} & 62.2 & 78.0 \\
            MinkUNet \citep{minkunet} &  73.3 & - \\
            Cylinder3d \citep{cylinder3d} & 76.1 & 77.2 \\
            SPVNAS \citep{spvnas} & 77.4 & - \\
            RPVNet \citep{rpvnet} & 77.6 & - \\
            RangeFormer \citep{rangeformer} & 78.1 & 80.1 \\
            SphereFormer \citep{sphereformer} & 78.4 & 81.9 \\
            OA-CNNs \citep{oacnns} & 78.9 & - \\
            PTv2 \citep{ptv2} & 80.2  & 82.6\\
            PTv3 \citep{ptv3} & 80.4 & 82.7 \\
            \midrule
            \baseline{\textsc{\methodname} (Ours)} & \baseline{\textbf{80.8}} & \baseline{\textbf{82.8}}  \\
            \bottomrule
        \end{tabular}}
    \end{minipage}
    \begin{minipage}{.44\linewidth}
        \centering
        \captionsetup{aboveskip=-0.01pt}  
        \captionsetup{belowskip=10pt}  
        \caption{Latency, parameters and accuracy\\ comparison on ScanNet.}
        \label{table5}
        \resizebox{0.98\linewidth}{!}{\begin{tabular}{c ccc}
        \toprule
        Method & Latency (ms) & Parameters (M) & mIoU \\
        \midrule
        Point Mamba \citep{pointmamba2} & 280 & 109.5 & 75.7 \\
        PTv2 \citep{ptv2} & 296 & \textbf{12.8} & 75.4 \\
        PTv3 \citep{ptv3} & 190 & 46.2 & 77.5 \\
        OctFormer \citep{octformer} & 133 & 44.0 & 75.7 \\
        OA-CNN \citep{oacnns} & 127 & 51.5 & 76.1\\
        \midrule
        \baseline{\textsc{\methodname} (Ours)} & \baseline{\textbf{110}} & \baseline{45.6} & \baseline{\textbf{78.0}} \\
        \bottomrule
    \end{tabular}}
        \end{minipage}
        \hfill
        \begin{minipage}{.55\linewidth}
        \centering
        \captionsetup{aboveskip=-0.01pt}  
        \captionsetup{belowskip=10pt}  
        \caption{Performance on ScanNet data efficient\\ benchmark.} 
        \label{table6}
        \resizebox{0.98\linewidth}{!}{\begin{tabular}{cccccccccc}\toprule
 \multirow{2}{*}{Method} &\multicolumn{4}{c}{Limited Reconstruction} &\multicolumn{4}{c}{Limited Annotation} \\\cmidrule(lr){2-5} \cmidrule(lr){6-9}
 &1\% &5\% &10\% &20\% &20 &50 &100 &200 \\
\midrule
MinkUNet \citep{minkunet} & 26.0 & 47.8 & 56.7 & 62.9 & 41.9 & 53.9 & 62.2 & 65.5 \\
PTv2 \citep{ptv2} & 24.8 & 48.1 & 59.8 & 66.3 & 58.4 & 66.1 & 70.3 & 71.2 \\
PTv3 \citep{ptv3} & 25.8 & 48.9 & 61.0 & 67.0 & 60.1 & 67.9 & 71.4 & 72.7 \\
\midrule
\baseline{\textsc{\methodname} (Ours)} & \baseline{\textbf{26.4}} & \baseline{\textbf{50.9}} & \baseline{\textbf{62.3}} & \baseline{\textbf{68.1}} & \baseline{\textbf{61.9}} & \baseline{\textbf{68.8}} & \baseline{\textbf{72.3}} & \baseline{\textbf{74.4}} \\
\bottomrule
\end{tabular}}
    \end{minipage}
    \vspace{-2mm}
\end{table}

In this section, we begin by briefly describing our implementation details (Sec.~\ref{sec5.1}). Then, we compare \textsc{\methodname} with state-of-the-art (SOTA) methods (Sec.~\ref{sec5.2}) and ablate our proposed method (Sec.~\ref{sec5.3}). Due to space limitation, detailed implementation details, more quantitative and qualitative results and additional ablations are presented in the Appendix \ref{appendix}.  

\subsection{Implementation Details}
\label{sec5.1}
\paragraph{Datasets.} We evaluate our proposed method on several indoor and outdoor semantic segmentation datasets using mean Intersection over Union (mIoU) metric. For indoor scenes, we use ScanNet \citep{scannet}, its extended version ScanNet200 \citep{scannet200}, and S3DIS \citep{s3dis}. For outdoor scenes, we employ nuScenes \citep{nuscenes}.

\paragraph{Training and Inference Details.} We follow all experimental settings of PTv3 \citep{ptv3} without any changes. For indoor segmentation, the number of epochs is 800, the learning rate is 0.006, and the weight decay is 0.05. For outdoor segmentation, the number of epochs is 50, the learning rate is 0.002, and the weight decay is 0.005. We train all our models using batch size of 12 and the AdamW optimizer. For efficiency evaluations, we use a single V100 with a batch size of 1.

\subsection{Main Results}
\label{sec5.2}
\paragraph{Indoor Semantic Segmentation.} Tab.~\ref{table3} compares \textsc{\methodname} with leading methods on the indoor ScanNet, ScanNet200, and S3DIS Area 5 cross-val datasets. \textsc{\methodname} achieves new SOTA results with mIoU scores of \textbf{78.0}, \textbf{36.0}, and \textbf{73.5} on ScanNet val, ScanNet200 val, and S3DIS Area 5 cross-val, respectively, outperforming the second-best method, PTv3, by \textbf{+0.5}, \textbf{+0.8}, and \textbf{+0.1} points.

\paragraph{Outdoor Semantic Segmentation.} Tab.~\ref{table4} compares \textsc{\methodname} with leading methods on the outdoor nuScenes val dataset. \textsc{\methodname} achieves new SOTA result with mIoU score of \textbf{80.8} on nuScenes val, surpassing the second-best method, PTv3, by \textbf{0.4} points.

\paragraph{Model Efficiency.} In Fig.~\ref{figure1}, we compare the average latency and memory usage of our model with multiple leading methods on the ScanNet val dataset. As shown in Fig.~\ref{figure1}(a) and Fig.~\ref{figure1}(b), \textsc{\methodname} exhibits better latency and memory consumption than many previous leading networks and achieves much higher performance. Remarkably, \textsc{\methodname} not only outperforms the previous top-performing method, PTv3, but is also $42.1\%$ faster and $5.53\times$ more memory efficient. \textsc{\methodname} particularly excels at ultra-long-range modeling, owing to its use of Mamba, which eliminates the quadratic complexity of attention typical in Transformer models. \textsc{\methodname}  also scales well to scenes containing a large number of points. As illustrated in  Fig.\ref{figure1}(d), \textsc{\methodname} also achieves $12\times$ reduction in FLOPs when handling scenes containing $131,072$ points. Note that Tab.~\ref{table5} gives the exact numbers corresponding to points in Fig.~\ref{figure1}(a) and Fig.~\ref{figure1}(b).

\paragraph{Data Efficiency.} In Tab.~\ref{table6}, we compare \textsc{\methodname} with leading methods on the ScanNet data efficiency benchmark, which evaluates models with limited reconstructions and annotations. The “Limited Reconstruction” setting uses a fraction of the available 3D reconstructions, while the “Limited Annotation” setting restricts the number of annotated points per scene. As shown, \textsc{\methodname} outperforms the second-best method, PTv3, by \textbf{0.6}, \textbf{2.0}, \textbf{1.3}, and \textbf{1.1} points at 1\%, 5\%, 10\%, and 20\% reconstructions, and by \textbf{1.8}, \textbf{0.9}, \textbf{0.9}, and \textbf{1.7} points at 20, 50, 100, and 200 annotations, respectively.
\begin{table}[t]
\vskip -0.1in
\captionsetup[subfloat]{labelformat=empty, labelfont=bf, textfont=normalfont, singlelinecheck=true}
\caption{\textbf{Ablation experiments on} on ScanNet for evaluating different design choices used for \textsc{\methodname}. The entries marked in \colorbox{baselinecolor}{gray} are the same, which specify the default settings.}
\label{table7}
\makebox[\textwidth][c]{\begin{minipage}{1\linewidth}
\centering
\vskip -0.1in
\subfloat[
(a) Effectiveness of Proposed Architecture
]{
\centering
\begin{minipage}{0.47\linewidth}
\vskip -0.05in
{\begin{center}
\tablestyle{3pt}{1.05}
\scriptsize{
\begin{tabular}{x{52}x{36}x{36}x{36}}
case & params (M) & latency (ms) & mIoU \\
\shline
Pure CNN & \textbf{41.6} & \textbf{80} & 73.2 \\
Pure Transformer  & 47.4 & 241 & 69.3 \\
Pure Mamba  & 48.4 & 126 & 70.7 \\
\textsc{\methodname} (Ours) & \baseline{45.6} & \baseline{110} & \baseline{\textbf{78.0}} \\
\end{tabular}}
\end{center}}\end{minipage}
}
\hspace{1em}
\subfloat[
(b) Effectiveness of CNN-Mamba Block
]{
\centering
\begin{minipage}{0.47\linewidth}
\vskip -0.05in
{\begin{center}
\tablestyle{3pt}{1.05}
\scriptsize{
\begin{tabular}{x{60}x{40}x{36}x{28}}
number of blocks & memory (GB) & latency (ms) & mIoU \\
\shline
22 & \baseline{\textbf{4.9}} & \baseline{\textbf{110}} & \baseline{\textbf{78.0}} \\
20  & 5.0 & 112 & 77.7 \\
12  & 5.6 & 117 & 77.5 \\
8  & 6.1 & 122 & 77.2 \\
4  & 10.3 & 132 & 77.3 \\
0 & 29.5 & 153 & 77.4 \\
\end{tabular}}
\end{center}}\end{minipage}
}
\hspace{1em}
\\
\vskip 0.1in
\subfloat[
(c) Effectiveness of Causal-Free Conv1D
]{
\centering
\begin{minipage}{0.47\linewidth}
\vskip -0.05in
{\begin{center}
\tablestyle{3pt}{1.05}
\scriptsize{
\begin{tabular}{x{72}x{48}}
case & mIoU \\
\shline
Causal Conv1D & 77.5 \\
Causal-Free Conv1D & \baseline{\textbf{78.0 \textcolor{forestgreen}{(+0.5)}}} \\
\\
\end{tabular}}
\end{center}}\end{minipage}
}
\hspace{1em}
\subfloat[
(d) Effectiveness of Bidirectional Strided SSM
]{
\centering
\begin{minipage}{0.47\linewidth}
\vskip -0.05in
{\begin{center}
\tablestyle{3pt}{1.05}
\scriptsize{
\begin{tabular}{x{96}x{48}}
case & mIoU \\
\shline
Standard SSM & 77.2 \\
Bidirectional SSM & 77.3 \textcolor{forestgreen}{(+0.1)} \\
Bidirectional Strided SSM & \baseline{\textbf{78.0 \textcolor{forestgreen}{(+0.8)}}} \\
\end{tabular}}
\end{center}}\end{minipage}
}
\hspace{1em}
\\
\vskip 0.1in
\subfloat[
(e) Optimal Stride for Bidirectional Strided SSM
]{
\centering
\begin{minipage}{0.47\linewidth}
\vskip -0.05in
{\begin{center}
\tablestyle{3pt}{1.05}
\scriptsize{\begin{tabular}{x{36}x{36}}
stride & mIoU \\
\shline
2 & \baseline{\textbf{78.0}} \\
4 & 77.7 \\
8 & 77.6 \\
16 & 77.5 \\
\end{tabular}}
\end{center}}\end{minipage}
}
\hspace{1em}
\subfloat[
(f) Compatibility of Our Proposed Modules
]{
\centering
\begin{minipage}{0.47\linewidth}
\vskip -0.05in
{\begin{center}
\tablestyle{3pt}{1.05}
\scriptsize{\begin{tabular}{lx{64}x{32}}
case & mIoU \\
\shline
baseline \scriptsize{(Pure CNN network)} & 73.2 \\
\quad + CNN-Mamba blocks & 76.9 \textcolor{forestgreen}{(+3.7)} \\
\quad \quad + Causal-Free Conv1D & 77.4 \textcolor{forestgreen}{(+4.2)} \\
\quad \quad \quad + Bidirectional Strided SSM & \baseline{78.0 \textcolor{forestgreen}{(+4.8)}} \\
\\
\end{tabular}}
\end{center}}\end{minipage}
}
\hspace{1em}
\vspace{-2em}
\end{minipage}}
\end{table}

\subsection{Ablation Experiments}
\label{sec5.3}
In this subsection, we conduct ablations for all components of our architecture using the ScanNet \citep{scannet} val dataset. All experimental settings follow the settings used in the main results.

\paragraph{Effectiveness of Our Proposed Architecture.}  Aside from the qualitative comparisons presented in Fig.~\ref{figure2} and Fig.~\ref{figure3}, we also quantitatively compare our proposed method with other model architectures. As shown in Tab.~\ref{table7}(a), \textsc{\methodname} requires fewer parameters and delivers superior performance.

\paragraph{Effectiveness of CNN-Mamba Block.} In Tab.~\ref{table7}(b), we progressively replace some of the CNN-Mamba blocks in \textsc{\methodname} with CNN-Transformer blocks. The results show that increasing the number of CNN-Transformer blocks always leads to higher latency, more memory usage and reduced performance. This confirms the effectiveness of our proposed CNN-Mamba block, which can efficiently and effectively integrate local and contextual modeling for point cloud segmentation.

\paragraph{Effectiveness of Causal-Free Mamba.} The original Mamba uses causal depthwise convolution to preprocess input tokens before passing them to the SSM module. While this makes sense for sequence modeling, its necessity for 3D vision tasks, which lack causality, is unclear. To investigate this, we experiment with replacing the causal convolution with normal convolution. As shown in Tab.~\ref{table7}(c), causal convolution results in a performance improvement of \textbf{+0.5} mIoU.

\paragraph{Effectiveness of Bidirectional Strided SSM.} In Tab.~\ref{table7}(d), we demonstrate the effectiveness of our proposed \textit{Bidirectional Strided SSM}. As shown, it outperforms both the standard SSM and its bidirectional variant, resulting in a performance improvement of \textbf{+0.8} mIoU. Additionally, we ablate the optimal stride for this module in Tab.~\ref{table7}(e), which shows that a stride of 2 yields the best performance.

\paragraph{Compatibility of Our Proposed Modules.} In addition to ablating our proposed modules individually, we perform an additive ablation study in Tab.~\ref{table7}(f) to demonstrate the compatibility of the enhancements. Incorporating CNN-Mamba blocks results in a significant improvement of \textbf{+3.7} mIoU. Further upgrades to the standard Mamba module, namely introducing causal-free convolutions and employing multi-directional scanning, provide additional gains of \textbf{+0.5} and \textbf{+0.6} mIoU, respectively.

\section{Related Work}

\noindent \textbf{Architectures for point cloud segmentation} fall into three main categories: point-based \citep{kpconv,pointnet,pointnet++,pointmlp}, voxel-based \citep{7353481,song2017semantic}, and projection-based \citep{chen2017multi,lang2019pointpillars,li2016vehicle,su2015multi}. Although they differ in pre-processing strategies, all these methods are designed with careful consideration of the unique characteristics of point clouds. They mainly differ in how they integrate local and contextual features and manage irregular point distributions. Recently, transformer-based models \citep{ptv3,superpointtransformer,swin3d,st} have emerged in this field, achieving higher accuracy but suffering from significant time and memory complexities relative to the size of the point cloud. To address this, more efficient attention mechanisms, such as vector attention \citep{zhao2020exploring}, grouped vector attention \citep{ptv2}, local window-based attention \citep{st}, and memory-efficient attention \citep{swin3d}, have been developed. However, these strategies approximate original attention, resulting in a loss of global modeling capability \citep{shen2021efficient}, which may impede performance on long range modeling task like point cloud segmentation. Our research examines the strengths and weaknesses of several widely-used architectures for point cloud segmentation and introduces a novel architecture that effectively integrates their best features, culminating in a new state-of-the-art model for point cloud segmentation.

\noindent \textbf{State space models (SSMs)} \citep{mamba,mambabyte,jamba} have emerged as a promising alternative to Transformers \citep{transformer} in natural language processing (NLP) for capturing long-range dependencies. Unlike Transformers that scale quadratically with sequence length, SSMs \citep{s4, s4nd, s5} achieve linear scaling during inference. The seminal Mamba model \citep{mamba} greatly improves the performance and efficiency of SSM by introducing input-specific parameterization and a scalable, hardware-optimized method, allowing it to outperform Transformers for the first time. Driven by the success of SSMs in NLP, recent works have also explored their application to visual tasks. S4ND \citep{s4nd} marks the introduction of SSM modules for processing visual data across 1D, 2D, and 3D domains. Following works, such as VMamba \citep{vmamba} and Vim \citep{vim}, address the directional sensitivity in SSMs with bi-directional and cross-scan mechanisms, showcasing the potential of SSMs to rival traditional CNN and Transformer models. The impressive performance of Mamba-based models has spurred further research into their applications for various vision tasks, such as image segmentation \citep{segmamba,swinumamba}, image synthesis \citep{diffussm}, graph modeling \citep{graphmamba}, and low-level vision tasks \citep{mambair}, underscoring their versatility and effectiveness. Despite some initial attempts to apply SSM architectures for point cloud segmentation, these models have failed to outperform existing architectures. In this work, we identify the shortcomings of SSM when applied to point cloud segmentation and propose simple solutions to significantly improve its performance in this domain.

\section{Conclusion}
In this work, we present a detailed analysis of existing network architectures for point cloud segmentation, highlighting their strengths and weaknesses. This evaluation provides valuable insights into designing more efficient and effective architectures for this task. Our findings emphasize the crucial role of \textit{spatial locality} and \textit{robust contextual understanding} in achieving strong performance. Specifically, we identify convolution and the Mamba module as essential components for efficient and accurate point cloud segmentation. Convolution provides spatial locality, while the Mamba module enhances the understanding of context. Additionally, we improve the standard Mamba module by removing the causality constraint and introducing \textit{Bidirectional Strided SSM}, which further enhances its ability to capture and utilize contextual information. Following these design principles and applying targeted optimizations, we introduce \textsc{\methodname}, a novel architecture that outperforms previous state-of-the-art models across multiple key benchmark datasets and efficiency metrics.

\newpage

\noindent \textbf{Acknowledgements.} This work is supported in part by the National Key R\&D Program of China under Grant 2021ZD0140407, the National Natural Science Foundation of China under Grants 62321005 and 62276150, and the THU-Bosch JCML.

\bibliography{iclr2025_conference}

\begin{thebibliography}{59}
\providecommand{\natexlab}[1]{#1}
\providecommand{\url}[1]{\texttt{#1}}
\expandafter\ifx\csname urlstyle\endcsname\relax
  \providecommand{\doi}[1]{doi: #1}\else
  \providecommand{\doi}{doi: \begingroup \urlstyle{rm}\Url}\fi

\bibitem[Armeni et~al.(2016)Armeni, Sener, Zamir, Jiang, Brilakis, Fischer, and Savarese]{s3dis}
Iro Armeni, Ozan Sener, Amir Zamir, Helen Jiang, Ioannis~K. Brilakis, Martin Fischer, and Silvio Savarese.
\newblock 3d semantic parsing of large-scale indoor spaces.
\newblock \emph{CVPR}, 2016.

\bibitem[Caesar et~al.(2020)Caesar, Bankiti, Lang, Vora, Liong, Xu, Krishnan, Pan, Baldan, and Beijbom]{nuscenes}
Holger Caesar, Varun Bankiti, Alex~H Lang, Sourabh Vora, Venice~Erin Liong, Qiang Xu, Anush Krishnan, Yu~Pan, Giancarlo Baldan, and Oscar Beijbom.
\newblock nuscenes: A multimodal dataset for autonomous driving.
\newblock In \emph{CVPR}, 2020.

\bibitem[Chen et~al.(2017)Chen, Ma, Wan, Li, and Xia]{chen2017multi}
Xiaozhi Chen, Huimin Ma, Ji~Wan, Bo~Li, and Tian Xia.
\newblock Multi-view 3d object detection network for autonomous driving.
\newblock In \emph{CVPR}, 2017.

\bibitem[Cheng et~al.(2021)Cheng, Razani, Taghavi, Li, and Liu]{af2s3net}
Ran Cheng, Ryan Razani, Ehsan Taghavi, Enxu Li, and Bingbing Liu.
\newblock 2-s3net: Attentive feature fusion with adaptive feature selection for sparse semantic segmentation network.
\newblock In \emph{CVPR}, 2021.

\bibitem[Cho et~al.(2014)Cho, van Merri{\"e}nboer, Bahdanau, and Bengio]{gru}
Kyunghyun Cho, Bart van Merri{\"e}nboer, Dzmitry Bahdanau, and Yoshua Bengio.
\newblock On the properties of neural machine translation: Encoder{--}decoder approaches.
\newblock In \emph{Proceedings of {SSST}-8, Eighth Workshop on Syntax, Semantics and Structure in Statistical Translation}, 2014.

\bibitem[Choy et~al.(2019)Choy, Gwak, and Savarese]{minkunet}
Christopher Choy, JunYoung Gwak, and Silvio Savarese.
\newblock 4d spatio-temporal convnets: Minkowski convolutional neural networks.
\newblock In \emph{CVPR}, 2019.

\bibitem[Contributors(2023)]{pointcept2023}
Pointcept Contributors.
\newblock Pointcept: A codebase for point cloud perception research.
\newblock \url{https://github.com/Pointcept/Pointcept}, 2023.

\bibitem[Dai et~al.(2017)Dai, Chang, Savva, Halber, Funkhouser, and Nie{\ss}ner]{scannet}
Angela Dai, Angel~X Chang, Manolis Savva, Maciej Halber, Thomas Funkhouser, and Matthias Nie{\ss}ner.
\newblock Scannet: Richly-annotated 3d reconstructions of indoor scenes.
\newblock In \emph{CVPR}, 2017.

\bibitem[Graham et~al.(2018)Graham, Engelcke, and Van Der~Maaten]{submanifold}
Benjamin Graham, Martin Engelcke, and Laurens Van Der~Maaten.
\newblock 3d semantic segmentation with submanifold sparse convolutional networks.
\newblock In \emph{CVPR}, 2018.

\bibitem[Gu \& Dao(2024)Gu and Dao]{mamba}
Albert Gu and Tri Dao.
\newblock Mamba: Linear-time sequence modeling with selective state spaces.
\newblock In \emph{COLM}, 2024.

\bibitem[Gu et~al.(2022)Gu, Goel, and R{\'e}]{s4}
Albert Gu, Karan Goel, and Christopher R{\'e}.
\newblock Efficiently modeling long sequences with structured state spaces.
\newblock In \emph{ICLR}, 2022.

\bibitem[Guo et~al.(2024)Guo, Li, Dai, Ouyang, Ren, and Xia]{mambair}
Hang Guo, Jinmin Li, Tao Dai, Zhihao Ouyang, Xudong Ren, and Shu-Tao Xia.
\newblock Mambair: A simple baseline for image restoration with state-space model.
\newblock In \emph{ECCV}, 2024.

\bibitem[Hochreiter \& Schmidhuber(1997)Hochreiter and Schmidhuber]{lstm}
Sepp Hochreiter and J{\"u}rgen Schmidhuber.
\newblock Long short-term memory.
\newblock \emph{Neural computation}, 1997.

\bibitem[Kong et~al.(2023)Kong, Liu, Chen, Ma, Zhu, Li, Hou, Qiao, and Liu]{rangeformer}
Lingdong Kong, Youquan Liu, Runnan Chen, Yuexin Ma, Xinge Zhu, Yikang Li, Yuenan Hou, Yu~Qiao, and Ziwei Liu.
\newblock Rethinking range view representation for lidar segmentation.
\newblock In \emph{ICCV}, 2023.

\bibitem[Lai et~al.(2022)Lai, Liu, Jiang, Wang, Zhao, Liu, Qi, and Jia]{st}
Xin Lai, Jianhui Liu, Li~Jiang, Liwei Wang, Hengshuang Zhao, Shu Liu, Xiaojuan Qi, and Jiaya Jia.
\newblock Stratified transformer for 3d point cloud segmentation.
\newblock In \emph{CVPR}, 2022.

\bibitem[Lai et~al.(2023)Lai, Chen, Lu, Liu, and Jia]{sphereformer}
Xin Lai, Yukang Chen, Fanbin Lu, Jianhui Liu, and Jiaya Jia.
\newblock Spherical transformer for lidar-based 3d recognition.
\newblock In \emph{CVPR}, 2023.

\bibitem[Lang et~al.(2019)Lang, Vora, Caesar, Zhou, Yang, and Beijbom]{lang2019pointpillars}
Alex~H Lang, Sourabh Vora, Holger Caesar, Lubing Zhou, Jiong Yang, and Oscar Beijbom.
\newblock Pointpillars: Fast encoders for object detection from point clouds.
\newblock In \emph{CVPR}, 2019.

\bibitem[LeCun et~al.(1989)LeCun, Boser, Denker, Henderson, Howard, Hubbard, and Jackel]{cnn}
Yann LeCun, Bernhard Boser, John Denker, Donnie Henderson, R.~Howard, Wayne Hubbard, and Lawrence Jackel.
\newblock Handwritten digit recognition with a back-propagation network.
\newblock In \emph{NeurIPS}, 1989.

\bibitem[Li et~al.(2016)Li, Zhang, and Xia]{li2016vehicle}
Bo~Li, Tianlei Zhang, and Tian Xia.
\newblock Vehicle detection from 3d lidar using fully convolutional network.
\newblock In \emph{RSS}, 2016.

\bibitem[Liang et~al.(2024)Liang, Zhou, Wang, Zhu, Xu, Zou, Ye, and Bai]{pointmamba}
Dingkang Liang, Xin Zhou, Xinyu Wang, Xingkui Zhu, Wei Xu, Zhikang Zou, Xiaoqing Ye, and Xiang Bai.
\newblock Pointmamba: A simple state space model for point cloud analysis.
\newblock In \emph{NeurIPS}, 2024.

\bibitem[Lieber et~al.(2024)Lieber, Lenz, Bata, Cohen, Osin, Dalmedigos, Safahi, Meirom, Belinkov, Shalev-Shwartz, et~al.]{jamba}
Opher Lieber, Barak Lenz, Hofit Bata, Gal Cohen, Jhonathan Osin, Itay Dalmedigos, Erez Safahi, Shaked Meirom, Yonatan Belinkov, Shai Shalev-Shwartz, et~al.
\newblock Jamba: A hybrid transformer-mamba language model.
\newblock \emph{arXiv:2403.19887}, 2024.

\bibitem[Liu et~al.(2024{\natexlab{a}})Liu, Yang, Zhou, Xi, Yu, Yu, Liang, Shi, Zhang, Zheng, et~al.]{swinumamba}
Jiarun Liu, Hao Yang, Hong-Yu Zhou, Yan Xi, Lequan Yu, Yizhou Yu, Yong Liang, Guangming Shi, Shaoting Zhang, Hairong Zheng, et~al.
\newblock Swin-umamba: Mamba-based unet with imagenet-based pretraining.
\newblock In \emph{MICCAI}, 2024{\natexlab{a}}.

\bibitem[Liu et~al.(2024{\natexlab{b}})Liu, Yu, Wang, Zheng, Deng, Ye, and Wang]{pointmamba2}
Jiuming Liu, Ruiji Yu, Yian Wang, Yu~Zheng, Tianchen Deng, Weicai Ye, and Hesheng Wang.
\newblock Point mamba: A novel point cloud backbone based on state space model with octree-based ordering strategy.
\newblock \emph{arXiv:2403.06467}, 2024{\natexlab{b}}.

\bibitem[Liu et~al.(2024{\natexlab{c}})Liu, Tian, Zhao, Yu, Xie, Wang, Ye, and Liu]{vmamba}
Yue Liu, Yunjie Tian, Yuzhong Zhao, Hongtian Yu, Lingxi Xie, Yaowei Wang, Qixiang Ye, and Yunfan Liu.
\newblock Vmamba: Visual state space model.
\newblock In \emph{NeurIPS}, 2024{\natexlab{c}}.

\bibitem[Ma et~al.(2022)Ma, Qin, You, Ran, and Fu]{pointmlp}
Xu~Ma, Can Qin, Haoxuan You, Haoxi Ran, and Yun Fu.
\newblock Rethinking network design and local geometry in point cloud: A simple residual mlp framework.
\newblock In \emph{ICLR}, 2022.

\bibitem[Maturana \& Scherer(2015)Maturana and Scherer]{7353481}
Daniel Maturana and Sebastian Scherer.
\newblock Voxnet: A 3d convolutional neural network for real-time object recognition.
\newblock In \emph{IROS}, 2015.

\bibitem[Morton(1966)]{morton}
G.~M. Morton.
\newblock A computer oriented geodetic data base; and a new technique in file sequencing.
\newblock 1966.

\bibitem[Nguyen et~al.(2022)Nguyen, Goel, Gu, Downs, Shah, Dao, Baccus, and R\'e]{s4nd}
Eric Nguyen, Karan Goel, Albert Gu, Gordon~W. Downs, Preey Shah, Tri Dao, Stephen~A. Baccus, and Christopher R\'e.
\newblock S4nd: Modeling images and videos as multidimensional signals using state spaces.
\newblock In \emph{NeurIPS}, 2022.

\bibitem[Pan et~al.(2021)Pan, Xia, Song, Li, and Huang]{pointformer}
Xuran Pan, Zhuofan Xia, Shiji Song, Li~Erran Li, and Gao Huang.
\newblock 3d object detection with pointformer.
\newblock In \emph{CVPR}, 2021.

\bibitem[Peano(1890)]{spacefilling}
Giuseppe Peano.
\newblock Sur une courbe, qui remplit toute une aire plane.
\newblock \emph{Mathematische Annalen}, 1890.

\bibitem[Peng et~al.(2024)Peng, Wu, Jiang, Chen, Zhao, Tian, and Jia]{oacnns}
Bohao Peng, Xiaoyang Wu, Li~Jiang, Yukang Chen, Hengshuang Zhao, Zhuotao Tian, and Jiaya Jia.
\newblock Oa-cnns: Omni-adaptive sparse cnns for 3d semantic segmentation.
\newblock In \emph{CVPR}, 2024.

\bibitem[Pi{\'o}ro et~al.(2024)Pi{\'o}ro, Ciebiera, Kr{\'o}l, Ludziejewski, and Jaszczur]{moemamba}
Maciej Pi{\'o}ro, Kamil Ciebiera, Krystian Kr{\'o}l, Jan Ludziejewski, and Sebastian Jaszczur.
\newblock Moe-mamba: Efficient selective state space models with mixture of experts.
\newblock \emph{arXiv:2401.04081}, 2024.

\bibitem[Qi et~al.(2017{\natexlab{a}})Qi, Su, Mo, and Guibas]{pointnet}
Charles~R Qi, Hao Su, Kaichun Mo, and Leonidas~J Guibas.
\newblock Pointnet: Deep learning on point sets for 3d classification and segmentation.
\newblock In \emph{CVPR}, 2017{\natexlab{a}}.

\bibitem[Qi et~al.(2017{\natexlab{b}})Qi, Yi, Su, and Guibas]{pointnet++}
Charles~Ruizhongtai Qi, Li~Yi, Hao Su, and Leonidas~J Guibas.
\newblock Pointnet++: Deep hierarchical feature learning on point sets in a metric space.
\newblock In \emph{NeurIPS}, 2017{\natexlab{b}}.

\bibitem[Qian et~al.(2022)Qian, Li, Peng, Mai, Hammoud, Elhoseiny, and Ghanem]{pointnext}
Guocheng Qian, Yuchen Li, Houwen Peng, Jinjie Mai, Hasan Hammoud, Mohamed Elhoseiny, and Bernard Ghanem.
\newblock Pointnext: Revisiting pointnet++ with improved training and scaling strategies.
\newblock In \emph{NeurIPS}, 2022.

\bibitem[Robert et~al.(2023)Robert, Raguet, and Landrieu]{superpointtransformer}
Damien Robert, Hugo Raguet, and Loic Landrieu.
\newblock Efficient 3d semantic segmentation with superpoint transformer.
\newblock In \emph{ICCV}, 2023.

\bibitem[Rozenberszki et~al.(2022)Rozenberszki, Litany, and Dai]{scannet200}
David Rozenberszki, Or~Litany, and Angela Dai.
\newblock Language-grounded indoor 3d semantic segmentation in the wild.
\newblock In \emph{ECCV}, 2022.

\bibitem[Shen et~al.(2021)Shen, Zhang, Zhao, Yi, and Li]{shen2021efficient}
Zhuoran Shen, Mingyuan Zhang, Haiyu Zhao, Shuai Yi, and Hongsheng Li.
\newblock Efficient attention: Attention with linear complexities.
\newblock In \emph{WACV}, 2021.

\bibitem[Smith et~al.(2023)Smith, Warrington, and Linderman]{s5}
Jimmy~T.H. Smith, Andrew Warrington, and Scott Linderman.
\newblock Simplified state space layers for sequence modeling.
\newblock In \emph{ICLR}, 2023.

\bibitem[Song et~al.(2017)Song, Yu, Zeng, Chang, Savva, and Funkhouser]{song2017semantic}
Shuran Song, Fisher Yu, Andy Zeng, Angel~X Chang, Manolis Savva, and Thomas Funkhouser.
\newblock Semantic scene completion from a single depth image.
\newblock In \emph{CVPR}, 2017.

\bibitem[Su et~al.(2015)Su, Maji, Kalogerakis, and Learned-Miller]{su2015multi}
Hang Su, Subhransu Maji, Evangelos Kalogerakis, and Erik Learned-Miller.
\newblock Multi-view convolutional neural networks for 3d shape recognition.
\newblock In \emph{ICCV}, 2015.

\bibitem[Tang et~al.(2020)Tang, Liu, Zhao, Lin, Lin, Wang, and Han]{spvnas}
Haotian Tang, Zhijian Liu, Shengyu Zhao, Yujun Lin, Ji~Lin, Hanrui Wang, and Song Han.
\newblock Searching efficient 3d architectures with sparse point-voxel convolution.
\newblock In \emph{ECCV}, 2020.

\bibitem[Thomas et~al.(2019)Thomas, Qi, Deschaud, Marcotegui, Goulette, and Guibas]{kpconv}
Hugues Thomas, Charles~R Qi, Jean-Emmanuel Deschaud, Beatriz Marcotegui, Fran{\c{c}}ois Goulette, and Leonidas~J Guibas.
\newblock Kpconv: Flexible and deformable convolution for point clouds.
\newblock In \emph{ICCV}, 2019.

\bibitem[Vaswani et~al.(2017)Vaswani, Shazeer, Parmar, Uszkoreit, Jones, Gomez, Kaiser, and Polosukhin]{transformer}
Ashish Vaswani, Noam Shazeer, Niki Parmar, Jakob Uszkoreit, Llion Jones, Aidan~N Gomez, {\L}ukasz Kaiser, and Illia Polosukhin.
\newblock Attention is all you need.
\newblock In \emph{NeurIPS}, 2017.

\bibitem[Wang et~al.(2024{\natexlab{a}})Wang, Tsepa, Ma, and Wang]{graphmamba}
Chloe Wang, Oleksii Tsepa, Jun Ma, and Bo~Wang.
\newblock Graph-mamba: Towards long-range graph sequence modeling with selective state spaces.
\newblock \emph{arXiv:2402.00789}, 2024{\natexlab{a}}.

\bibitem[Wang et~al.(2024{\natexlab{b}})Wang, Gangavarapu, Yan, and Rush]{mambabyte}
Junxiong Wang, Tushaar Gangavarapu, Jing~Nathan Yan, and Alexander~M Rush.
\newblock Mambabyte: Token-free selective state space model.
\newblock In \emph{COLM}, 2024{\natexlab{b}}.

\bibitem[Wang(2023)]{octformer}
Peng-Shuai Wang.
\newblock Octformer: Octree-based transformers for 3d point clouds.
\newblock \emph{SIGGRAPH}, 2023.

\bibitem[Wu et~al.(2022)Wu, Lao, Jiang, Liu, and Zhao]{ptv2}
Xiaoyang Wu, Yixing Lao, Li~Jiang, Xihui Liu, and Hengshuang Zhao.
\newblock Point transformer v2: Grouped vector attention and partition-based pooling.
\newblock In \emph{NeurIPS}, 2022.

\bibitem[Wu et~al.(2023)Wu, Wen, Liu, and Zhao]{sparseunet}
Xiaoyang Wu, Xin Wen, Xihui Liu, and Hengshuang Zhao.
\newblock Masked scene contrast: A scalable framework for unsupervised 3d representation learning.
\newblock In \emph{CVPR}, 2023.

\bibitem[Wu et~al.(2024)Wu, Jiang, Wang, Liu, Liu, Qiao, Ouyang, He, and Zhao]{ptv3}
Xiaoyang Wu, Li~Jiang, Peng-Shuai Wang, Zhijian Liu, Xihui Liu, Yu~Qiao, Wanli Ouyang, Tong He, and Hengshuang Zhao.
\newblock Point transformer v3: Simpler, faster, stronger.
\newblock In \emph{CVPR}, 2024.

\bibitem[Xing et~al.(2024)Xing, Ye, Yang, Liu, and Zhu]{segmamba}
Zhaohu Xing, Tian Ye, Yijun Yang, Guang Liu, and Lei Zhu.
\newblock Segmamba: Long-range sequential modeling mamba for 3d medical image segmentation.
\newblock In \emph{MICCAI}, 2024.

\bibitem[Xu et~al.(2021)Xu, Zhang, Dou, Zhu, Sun, and Pu]{rpvnet}
Jianyun Xu, Ruixiang Zhang, Jian Dou, Yushi Zhu, Jie Sun, and Shiliang Pu.
\newblock Rpvnet: A deep and efficient range-point-voxel fusion network for lidar point cloud segmentation.
\newblock In \emph{ICCV}, 2021.

\bibitem[Yan et~al.(2024)Yan, Gu, and Rush]{diffussm}
Jing~Nathan Yan, Jiatao Gu, and Alexander~M Rush.
\newblock Diffusion models without attention.
\newblock In \emph{CVPR}, 2024.

\bibitem[Yang et~al.(2023)Yang, Guo, Xiong, Liu, Pan, Wang, Tong, and Guo]{swin3d}
Yu-Qi Yang, Yu-Xiao Guo, Jian-Yu Xiong, Yang Liu, Hao Pan, Peng-Shuai Wang, Xin Tong, and Baining Guo.
\newblock Swin3d: A pretrained transformer backbone for 3d indoor scene understanding.
\newblock \emph{CVMJ}, 2023.

\bibitem[Zhang et~al.(2024)Zhang, Li, Yuan, Ji, and Yan]{pcm}
Tao Zhang, Xiangtai Li, Haobo Yuan, Shunping Ji, and Shuicheng Yan.
\newblock Point could mamba: Point cloud learning via state space model.
\newblock \emph{arXiv:2403.00762}, 2024.

\bibitem[Zhao et~al.(2020)Zhao, Jia, and Koltun]{zhao2020exploring}
Hengshuang Zhao, Jiaya Jia, and Vladlen Koltun.
\newblock Exploring self-attention for image recognition.
\newblock In \emph{CVPR}, 2020.

\bibitem[Zhao et~al.(2021)Zhao, Jiang, Jia, Torr, and Koltun]{ptv1}
Hengshuang Zhao, Li~Jiang, Jiaya Jia, Philip~HS Torr, and Vladlen Koltun.
\newblock Point transformer.
\newblock In \emph{NeurIPS}, 2021.

\bibitem[Zhu et~al.(2024)Zhu, Liao, Zhang, Wang, Liu, and Wang]{vim}
Lianghui Zhu, Bencheng Liao, Qian Zhang, Xinlong Wang, Wenyu Liu, and Xinggang Wang.
\newblock Vision mamba: Efficient visual representation learning with bidirectional state space model.
\newblock In \emph{ICML}, 2024.

\bibitem[Zhu et~al.(2021)Zhu, Zhou, Wang, Hong, Ma, Li, Li, and Lin]{cylinder3d}
Xinge Zhu, Hui Zhou, Tai Wang, Fangzhou Hong, Yuexin Ma, Wei Li, Hongsheng Li, and Dahua Lin.
\newblock Cylindrical and asymmetrical 3d convolution networks for lidar segmentation.
\newblock In \emph{CVPR}, 2021.

\end{thebibliography}
\bibliographystyle{iclr2025_conference}

\newpage
\appendix

\section{Appendix}
\label{appendix}
For a thorough understanding of our proposed \textsc{\methodname}, we have compiled a detailed Appendix. The table of contents below offers a quick overview and will guide to
specific sections of interest.

\localtableofcontents

\subsection{Implementation Details.}
\label{a1}
We implement our method using the Pointcept \citep{pointcept2023} codebase. Detailed specifications of our implementation are provided in this section.

\begin{table}[h]
  \vskip -0.15in
    \begin{minipage}{.49\linewidth}
        \small
        \captionsetup{aboveskip=0pt}  
        \vskip 0.1in
        \caption{Indoor sem. seg. settings.}
        \label{table8}
        \centering
        \begin{tabular}{cc}
            \toprule 
            Config &Value \\
            \midrule
            optimizer &AdamW  \\
            scheduler &Cosine  \\
            criteria &CrossEntropy (1)  \\
            &Lovasz (1)  \\
            learning rate &6e-3  \\
            block lr scaler &0.1  \\
            weight decay &5e-2  \\
            batch size &12 \\
            datasets & ScanNet / S3DIS  \\
            warmup epochs &40  \\
            epochs &800  \\
            \bottomrule
            \end{tabular}
    \end{minipage}
    \hfill
    \begin{minipage}{.49\linewidth}
        \small
        \captionsetup{aboveskip=0pt}  
        \vskip 0.1in
        \caption{Outdoor sem. seg. settings.} 
        \label{table9}
        \centering
        \begin{tabular}{cc}
            \toprule 
            Config &Value \\
            \midrule
            optimizer &AdamW  \\
            scheduler &Cosine  \\
            criteria &CrossEntropy (1)  \\
            &Lovasz (1)  \\
            learning rate &2e-3  \\
            block lr scaler &0.1  \\
            weight decay &5e-3  \\
            batch size &12 \\
            datasets & nuScenes  \\
            warmup epochs &2  \\
            epochs & 50  \\
            \bottomrule
            \end{tabular}
    \end{minipage}
    \vspace{-2mm}
\end{table}

\paragraph{Training Settings.} The experimental settings for indoor and outdoor semantic segmentation are outlined in Tab.~\ref{table8} and Tab.~\ref{table9}. The numbers in brackets indicate the relative weight assigned to each criterion in the loss. The main differences between indoor and outdoor settings are in the learning rate, weight decay, warmup epochs and training epochs used.

\begin{table*}[h]
\vskip -0.05in
\centering
\captionsetup{aboveskip=-0.01pt}  
\caption{Model settings.}
\label{table10}
\resizebox{0.55\linewidth}{!}{\begin{tabular}{cc}
\toprule
Config & Value \\
\midrule
positional encoding & None \\
embedding depth & 2 \\
embedding channels & 32 \\
no. of layers in Local Perceiver & 1 \\
no. of layers in Channel Modulator & 2 \\
encoder depth & [2, 2, 6, 2] \\
encoder channels & [64, 128, 256, 512] \\
encoder num heads & [4, 8, 16, 32] \\
decoder depth & [1, 1, 1, 1] \\
decoder channels &[64, 64, 128, 256] \\
decoder num heads & [4, 4, 8, 16] \\
down stride &[$\times$2, $\times$2, $\times$2, $\times$2] \\
drop path & 0.3 \\
\bottomrule
\end{tabular}}
\end{table*}

\paragraph{Model Settings} Detailed model configurations of our \textsc{\methodname} are listed in Tab.~\ref{table10}.

\begin{table*}[h]

\centering
\captionsetup{aboveskip=-0.01pt}  
\caption{Data augmentations.}
\label{table11}
\begin{tabular}{llcc}\toprule
Augmentations &Parameters &Indoor &Outdoor \\\midrule
random dropout &dropout ratio: 0.2, p: 0.2 &\checkmark &- \\
random rotate &axis: z, angle: [-1, 1], p: 0.5 &\checkmark &\checkmark \\
&axis: x, angle: [-1 / 64, 1 / 64], p: 0.5 &\checkmark &- \\
&axis: y, angle: [-1 / 64, 1 / 64], p: 0.5 &\checkmark &- \\
random scale &scale: [0.9, 1.1] &\checkmark &\checkmark \\
random flip &p: 0.5 &\checkmark &\checkmark \\
random jitter &sigma: 0.005, clip: 0.02 &\checkmark &\checkmark \\
elastic distort & params: [[0.2, 0.4], [0.8, 1.6]] &\checkmark &- \\
auto contrast &p: 0.2 &\checkmark &- \\
color jitter &std: 0.05; p: 0.95 &\checkmark &- \\
grid sampling &grid size: 0.02 (indoor), 0.05 (outdoor) &\checkmark &\checkmark \\
sphere crop & max points: 102400 &\checkmark &- \\
normalize color &p: 1 &\checkmark &- \\
\bottomrule
\end{tabular}
\end{table*}

\newpage
\paragraph{Data Augmentation.} Data augmentations used for training \textsc{\methodname} are detailed in Tab.~\ref{table11}. 

\subsection{Additional Quantitative Results}

\begin{table*}[h]
    \centering
    \captionsetup{aboveskip=-0.01pt}  
    \caption{Results on ScanNet200 for semantic segmentation. }
    \label{table12}
    \tablestyle{6.9pt}{1.05}
    \begin{tabular}{c|cccc}
        \toprule
        \multirow{2}{*}{Method}&\multicolumn{4}{c}{Val}\\
        &Head&Comm.&Tail& All\\
        \midrule
        MinkowskiNet \citep{minkunet} & 48.3 & 19.1 & 7.9 & 25.1\\
        SparseUNet \citep{sparseunet} & - & - & - & 28.8 \\
        LGround \citep{scannet200} & 51.5 & 22.7 & 12.5 & 28.9 \\
        PTv2 \citep{ptv2} & - & - & - & 29.3 \\
        OA-CNNs \citep{oacnns} & 51.3 & 28.0 & 17.7 & 32.3 \\
        PTv3 \citep{ptv3} & 56.5 & 30.1 & 19.3 & 35.2 \\
        \midrule
        \baseline{\textsc{\methodname}} & \baseline{\textbf{56.6}} & \baseline{\textbf{30.7}} & \baseline{\textbf{20.7}} & \baseline{\textbf{36.0}}  \\
        \bottomrule
    \end{tabular}
\end{table*}
\paragraph{Class Imbalance Analysis on ScanNet200 \citep{scannet200}.} In Tab.~\ref{table12}, we compare \textsc{\methodname} with other leading methods on the \textit{head}, \textit{common}, and \textit{tail} subsets of the ScanNet200 validation benchmark. This comparison offers a more nuanced understanding of performance across different levels of class imbalance. As shown, \textsc{\methodname} achieves improvements of +0.2, +0.6, and +1.4 mIoU on these three subsets, respectively. Notably, it performs much better on the most challenging \textit{tail} subset, indicating its potential in handling long-tail distributions.

\begin{table*}[h]
\centering
\captionsetup{aboveskip=-0.01pt}   
\caption{Additional Evidences of Mamba's Importance.}
\label{table13}
\tablestyle{6.9pt}{1.05}
\begin{tabular}{lcc}
    \toprule
    Case & Params (M)$\downarrow$ & mIoU$\uparrow$  \\
    \midrule
    \textsc{\methodname} & \baseline{45.6} & \baseline{78.0}  \\
    \quad Replace Mamba with Attention & 42.7 & 77.4  \\
    \quad Remove Mamba & 38.2 & 73.5 \\
    \quad \quad Use one additional sparse conv. layer &69.0& 75.7 \\
    \quad \quad Use one additional block in each stage & 52.7 & 74.7 \\
    \quad \quad Increase input channel size from 32 to 36 & 48.3 & 74.2 \\
    \bottomrule
\end{tabular}
\end{table*}
\paragraph{Additional Evidences of Mamba's Importance.} To emphasize Mamba's importance, we also test alternative methods to scale up the models without it. The results in Tab.~\ref{table13} show that none of the \textsc{\methodname} variants without Mamba outperform the original configuration, clearly demonstrating its critical role in providing contextual modeling for point cloud segmentation.

\newpage
\subsection{Additional Qualitative Results}

\begin{figure*}[h]
\vskip -0.1in
  \centering
  \includegraphics[width=\linewidth]{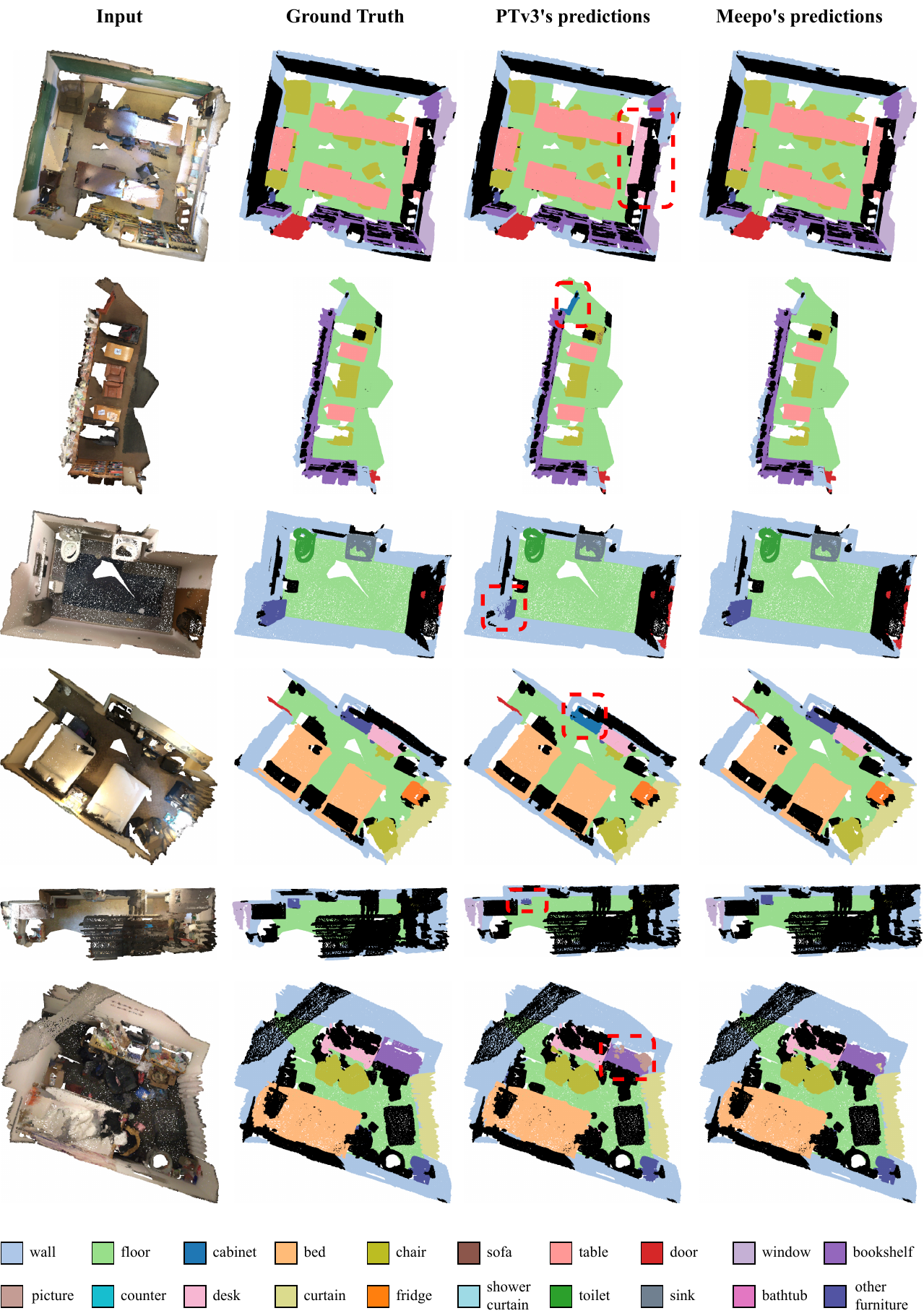}
  \captionsetup{belowskip=-10pt} 
  \caption{Comparison between \textsc{\methodname}'s and PTv3's \citep{ptv3} predictions. Black color are unlabelled points. Red boxes with dash-dotted lines are wrong predictions by PTv3.}
  \label{figure7}
\end{figure*}

\subsection{Limitations}

While our work has significantly advanced the performance of point cloud segmentation models, many challenges and opportunities for improvement remain. For instance, although \textsc{\methodname} exhibits notable efficiency gains, there is still room for further optimization to enhance its computational and memory efficiency. Additionally, segmentation quality, especially in handling fine details and complex geometries within point clouds, can be further improved. Enhancing the model's ability to accurately segment objects in diverse and densely populated scenes is another critical area for future research. Moreover, the potential of pretraining through self-supervised learning is unexplored in this work. Leveraging large-scale unlabeled point cloud data for pretraining could help the model learn more robust and generalizable features, ultimately boosting performance across various tasks and datasets. Future work should explore and integrate self-supervised learning techniques to harness this potential fully. Addressing these challenges will ensure that Mamba-based architectures continue to evolve and set new benchmarks in the field of point cloud segmentation.

\subsection{Broader Impacts}
\paragraph{Accessibility and Resource Efficiency.} By demonstrating that our proposed method, \textsc{\methodname} can achieve superior performance with reduced latency and memory consumption, this research promotes the development of more accessible and resource-efficient machine learning models. This is particularly important for applications in resource-constrained environments, such as mobile and embedded systems, where computational power and memory are limited. As a result, more organizations and developers can leverage advanced point cloud segmentation techniques without requiring extensive computational resources.

\paragraph{Environmental Impact.} The reduction in computational cost and memory usage directly translates to lower energy consumption. Given the growing concern over the environmental impact of large-scale machine learning models, \textsc{\methodname}'s efficiency can contribute to more sustainable AI practices. By minimizing the energy required for training and inference, this work aligns with global efforts to reduce the carbon footprint of technology.

\subsection{Compute Resources}
We run all experiments on a cluster with a mix of RTX3090, RTX4090 or 32GB V100 and 80GB A100 GPUs.

\subsection{Reproducibility}
Our main results can be fully reproduced by running the training and evaluation scripts given in the attached code.

\end{document}